\documentclass[10pt]{article}
\usepackage[preprint]{tmlr}

\usepackage{amsmath,amsfonts,bm}

\def\eqref#1{equation~\ref{#1}}

\def\1{\bm{1}}

\DeclareMathAlphabet{\mathsfit}{\encodingdefault}{\sfdefault}{m}{sl}
\SetMathAlphabet{\mathsfit}{bold}{\encodingdefault}{\sfdefault}{bx}{n}

\usepackage{float}
\usepackage{hyperref}
\usepackage{url}

\title{Test-Time Scaling in Reasoning LLMs: Inference Regimes, Evaluation, and Reproducibility}

\author{\name Mohsen Hariri \quad Weicong Chen \quad Nahal Shahini \quad Vikash Singh \quad Kai Ye \\
      \name Amirhossein Samandar \quad Debargha Ganguly \quad Sreehari Sankar \quad Yanyan Zhang \\
      \name Shouren Wang \quad Jerry Peng \quad Biyao Zhang \quad Michael Hinczewski \quad Vipin Chaudhary \\
      \email \{mohsen.hariri,weicong,nxs814,vikash,kxy406,axs2935,debargha,sxs2284,yxz3106,sxw992,jxp1146,bxz297,\\
      \email mxh605,vipin\}@case.edu \\
      \addr Case Western Reserve University}

\usepackage{custom}

\hypersetup{
  hidelinks,
  pdftitle={Test-Time Scaling in Reasoning LLMs: Inference Regimes, Evaluation, and Reproducibility},
  pdfauthor={Mohsen Hariri, Weicong Chen, Nahal Shahini, Vikash Singh, Kai Ye,
    Amirhossein Samandar, Debargha Ganguly, Sreehari Sankar, Yanyan Zhang,
    Shouren Wang, Jerry Peng, Biyao Zhang, Michael Hinczewski, Vipin Chaudhary}
}

\begin{document}
\maketitle

\begin{abstract}
Large language models can solve harder reasoning problems with more inference-time compute.
The term \emph{test-time scaling}, however, covers several inference algorithms: extending deliberation along one trajectory, sampling completed candidates and aggregating them by voting or verification, and searching over partial states. These algorithms differ in statistical structure, compute requirements, and failure modes. Treating them as interchangeable under a scalar ``budget,'' or reporting accuracy without specifying the inference protocol, makes results difficult to compare across studies. We study test-time scaling along three axes. First, we formalize it as budgeted inference over the implicit prefix tree of an autoregressive model and distinguish single-trajectory sequential scaling, leaf-level scaling with terminal reduction, and prefix-level scaling. Second, we treat the full inference system as the evaluated object and separate end-to-end performance from candidate-bank diagnostics. We introduce an evaluation profile whose coordinates and simple functionals recover or bound common repeated-sampling metrics, and require compute accounting and uncertainty estimates that match the protocol. Third, we distinguish exact replay from distributional reproducibility and state the requirements for each. We also organize open-weight reasoning models by model-side and interface mechanisms. Our empirical study covers broad knowledge, symbolic reasoning, and competition mathematics, and we publicly release 1,403,520 sampled model attempts.\footnote{For the API list, see Appendix~\ref{app:aggregation_software}; for the release data, see Appendix~\ref{app:dataset}.}
\end{abstract}
\begin{center}
\small
\scorioicon\enspace\softwareurl
\\[0.35em]
\textbf{Datasets:}\enspace
\huggingfaceicon~\datasettraces
\quad
\huggingfaceicon~\datasetlite
\quad
\huggingfaceicon~\datasetmath
\quad
\huggingfaceicon~\datasetgpqa
\end{center}

\section{Introduction}

Large language models (LLMs) often perform better on reasoning tasks when they generate intermediate steps and use more computation at inference time. Early work on chain-of-thought prompting and self-consistency showed gains on arithmetic, symbolic, and commonsense tasks from generating multi-step solutions and aggregating samples~\citep{wei2022cot,wang2022selfconsistency}. More recent systems make deliberate reasoning a training and interface objective rather than relying on prompting alone: DeepSeek-R1 uses reinforcement learning to elicit long-form reasoning~\citep{guo2025deepseek}; Phi-4-reasoning uses curated supervised traces, while Phi-4-reasoning-plus adds outcome-based reinforcement learning~\citep{abdin2025phi4reasoning}; Qwen3 exposes explicit ``thinking'' modes and reasoning-budget controls~\citep{qwen3technicalreport}; and the gpt-oss release provides open-weight reasoning models post-trained with chain-of-thought reinforcement learning~\citep{openai2025gptoss120bgptoss20bmodel}. Reasoning performance therefore depends on how computation is allocated at test time as well as on pretraining and post-training.

The phrase \emph{test-time scaling} describes several distinct inference algorithms. Some methods allocate extra compute to a single trajectory by forcing longer deliberation or intervening adaptively during generation~\citep{muennighoff2025s1}. Others sample completed candidates and reduce them by voting, reranking, verifier-based selection, or minimum-Bayes-risk decoding~\citep{wang2022selfconsistency,cobbe2021verifiers,freitag2022neuralmbr}. Still others search over partial states, as in Tree-of-Thoughts, RAP, and value-guided decoding~\citep{yao2023tot,hao2023rap,yu2024ovm}. These regimes differ in statistical structure, induced proposal distributions, compute requirements, and failure modes. Treating them as interchangeable under a scalar ``budget'' obscures the procedure being evaluated and makes cross-paper comparisons difficult to interpret~\citep{welleck2024inference_time_algos,snell2025computeoptimal}.

Evaluating these methods requires more than benchmarking a base model in isolation. Measured performance depends on the full inference protocol: the prompt template, decoder, search controller, reducer, verifier or judge, stopping rule, and numerical settings. Best-of-$N$ and verifier-guided reranking can improve accuracy, but selection can overoptimize imperfect proxy scores as the candidate set grows~\citep{gao2023scaling,huang2025bestofn,khalaf2025rewardhacking}. LLM judges can evaluate open-ended outputs, but prior work documents position and verbosity biases, and their reliability depends on the task and procedure in test-time scaling settings~\citep{zheng2023judge,zhou2025jetts}. Even with fixed prompts and methods, stochastic decoding and implementation details can change benchmark estimates~\citep{miller2024adding,blackwell2024towards,ye2024benchmarking,liu2025quant}. Studies of reasoning under test-time scaling must define the algorithm being evaluated and match the evaluation to the deployed inference protocol. They should also report model training, inference controls, and uncertainty with enough procedural detail for independent reruns.

\begin{figure}[t]
    \centering
    \includegraphics[width=\linewidth]{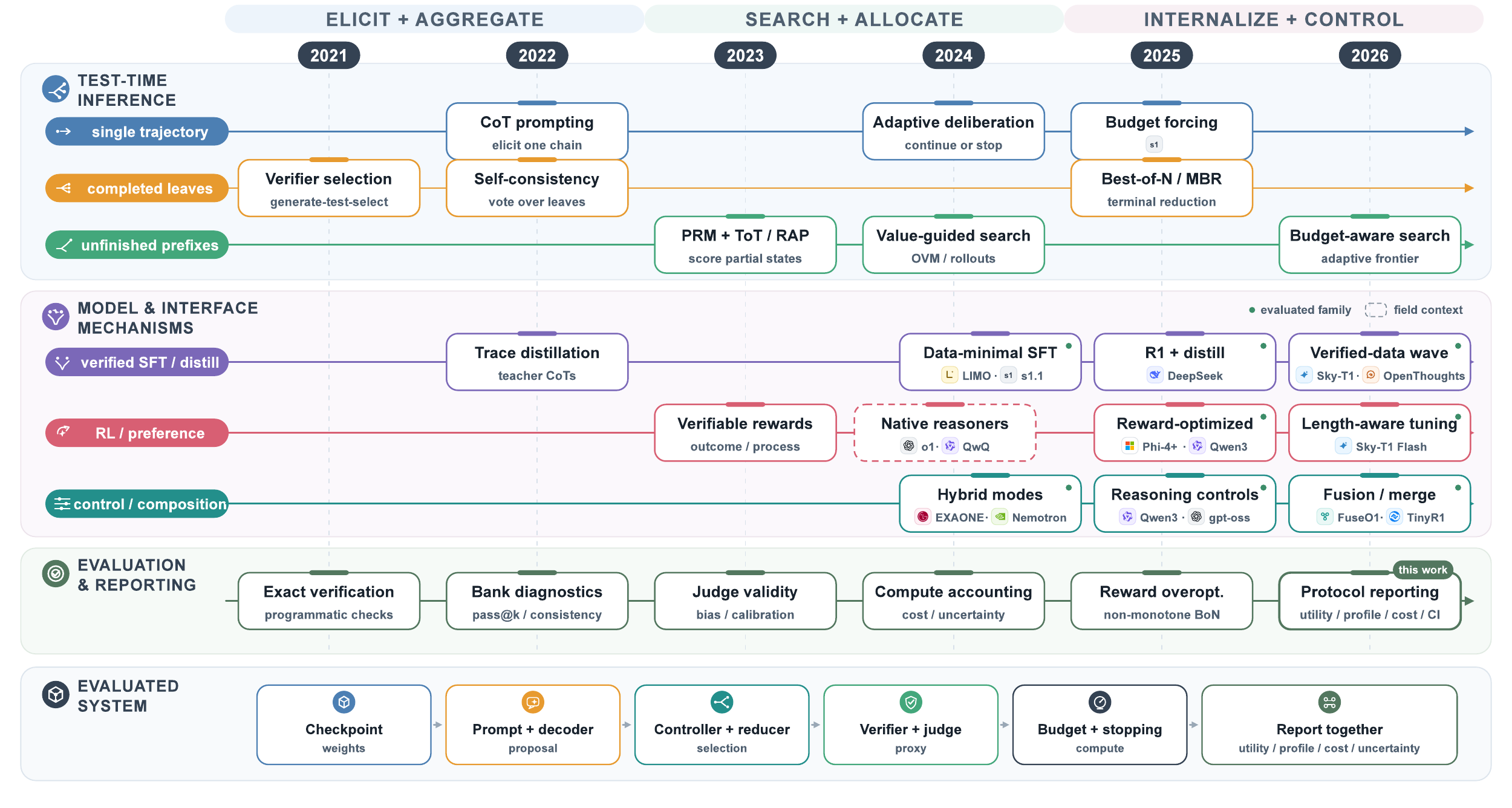}
    \caption{A selective chronology of reasoning systems from elicited chains to budgeted inference. The top band traces single-trajectory elicitation, leaf-level sampling with terminal reduction, prefix-level search, and budget-aware control (\Cref{sec:form:setup_taxonomy}). The middle band groups model and interface mechanisms into verified distillation, reward optimization, reasoning controls, and parameter-space composition (Appendices~\ref{app:reasoning_llms} and~\ref{app:prompt_templates}). The lower bands show evaluation and reproducibility requirements (\Cref{sec:eval}) and identify the evaluated system: checkpoint, prompt, decoder, controller or reducer, verifier or judge, budget, and stopping rule. Utility, candidate-bank profile, cost, and uncertainty are reported together. Green dots mark families represented in the experimental roster; dashed boxes denote contextual milestones. Era labels indicate shifts in emphasis rather than mutually exclusive periods.}
    \label{fig:reasoning-systems-chronicle}
\end{figure}

Comparisons of open-weight reasoning models require system-level evaluation because their training and inference mechanisms vary. Recent releases include reinforcement-learning-first reasoning models~\citep{guo2025deepseek}, data-centric and verification-focused SFT pipelines~\citep{ye2025limo,guha2025openthoughts}, open reproductions built from released trace corpora and training code~\citep{openr1_repo,mixture_of_thoughts}, checkpoints with thinking and non-thinking modes~\citep{qwen3technicalreport}, and model-fusion systems that combine reasoning experts~\citep{fuseo1_32b_preview,fuseo1_flash_32b_preview}. This variation confounds checkpoint comparisons because two checkpoints may differ at once in their supervision signals, reasoning styles, compute budgets, and aggregation protocols. Comparisons should separate these axes and treat reasoning as a property of the full inference system rather than of model weights alone. The chronology in \Cref{fig:reasoning-systems-chronicle} places these developments alongside the components that determine system performance: checkpoint, interface, controller or reducer, evaluator, stopping rule, and budget.

We formalize test-time scaling as a family of budgeted inference algorithms over the implicit prefix tree of an autoregressive model. For sampled inference, the formalization separates evidence computed for each trace from decisions over a completed pool. It treats causal stopping separately because stopping changes generation cost. Shared-bank comparisons isolate aggregation, whereas end-to-end evaluation measures the full inference system. We report candidate-bank quality with repeated-sampling diagnostics that capture the trade-off between discovering a correct candidate and producing correct candidates consistently. We apply the framework to open-weight reasoning models on benchmarks in broad knowledge, symbolic reasoning, and competition mathematics.

The paper makes four contributions:
\begin{itemize}
    \item \textbf{Formalizing test-time scaling.} We model test-time scaling as budgeted inference over the implicit prefix tree and distinguish single-trajectory sequential scaling, leaf-level scaling, and prefix-level scaling.
    \item \textbf{Establishing evaluation principles for test-time scaling.} We treat the base model, prompting, decoder, inference-time evidence, search or aggregation rule, stopping controller, evaluator, and budget as one inference system. We distinguish shared-bank aggregation comparisons from end-to-end evaluation and post-hoc selection from causal stopping. The evaluation principles also require compute reporting on the relevant axes and uncertainty estimates matched to the protocol.
    \item \textbf{Providing a model-centric view of the open reasoning ecosystem.} We organize and compare representative open-weight reasoning models by model-side and interface mechanisms: verified SFT and distillation, reinforcement learning or preference optimization, inference-time control, and parameter-space fusion.
    \item \textbf{Releasing a large-scale reasoning-trace resource.} We release 1,403,520 sampled attempts across competition mathematics and broad graduate-level knowledge for reasoning-behavior analysis, evaluator training or calibration, and reproducibility studies.
\end{itemize}

\section{Formalizing Test-Time Scaling}
\label{sec:form:tts}

Test-time scaling comprises a family of \emph{budgeted inference algorithms} built on a fixed autoregressive model. Repeated sampling with terminal reduction, adaptive single-trajectory deliberation, and search over partial reasoning states allocate additional inference-time computation in different ways~\citep{welleck2024inference_time_algos,snell2025computeoptimal,muennighoff2025s1,wu2024inference}. All operate over the implicit prefix tree induced by generation: compute may be spent along one trajectory, across completed leaves, or over internal prefixes before completion. These choices define three structural regimes.

\subsection{Problem setup and three-regime taxonomy}
\label{sec:form:setup_taxonomy}

Let $x\in\mathcal{X}$ be an input prompt and let $p_\theta(\cdot\mid x)$ be an autoregressive language model. The finite prefixes reachable from $x$ define an implicit rooted tree $\mathcal{T}(x)$; we write $z \preceq y$ when prefix $z$ lies on the path to completed generation $y$, and $\mathcal{L}(x)$ for the set of terminal leaves~\citep{welleck2024inference_time_algos}. A local generation policy $\pi$, such as greedy decoding, temperature sampling, or nucleus sampling, induces a proposal distribution $q_\pi(y\mid x)$ over completed leaves. This proposal may differ from the raw model distribution because decoding can truncate, renormalize, or otherwise transform token probabilities. We treat EOS, a declared length cap, and a causal stop issued by a controller as terminal actions; a stopped output that cannot be parsed remains a terminal leaf with invalid answer $a=\bot$.

A completed generation $y$ may contain both intermediate reasoning and a task answer. We write $\mathrm{Parse}(y)=(r,a)$, where $r$ is the reasoning trace and $a$ is the extracted answer. When only part of the output can be checked programmatically, we further write $\Psi(y)=(d,s)$, where $d$ is the deterministically verifiable component and $s$ is the remainder. Depending on the task, $d$ may be the final answer string, an executable program, an action sequence, or any other artifact that admits task-specific checking.

Our object of interest is a family of budgeted inference algorithms $\{\mathcal{A}_B\}_{B\in\mathcal{B}}$, where $\mathcal{B}\subseteq\mathbb{R}_{\ge 0}$ is the set of allowed budgets. Each family defines an additive cost function $c$ with a declared unit; resources that are not commensurate are reported separately. Algorithm $\mathcal{A}_B$ may adaptively interleave primitive operations $o_t$ such as token generation, prefix expansion, verifier calls, judge calls, or terminal reduction, subject to $\sum_t c(o_t)\le B$. Let $\mathcal{O}(x)=\mathcal{L}(x)\sqcup\mathcal{A}_{\mathrm{ans}}(x)$ be the disjoint union of completed leaves and answer-valued outputs. The algorithm returns $\hat{o}_B(x)\in\mathcal{O}(x)$, written $\hat{y}_B(x)$ or $\hat{a}_B(x)$ according to its type, and $U_x:\mathcal{O}(x)\to\mathbb{R}$ denotes task utility. In reasoning benchmarks $U_x$ is often exact correctness, but the same abstraction also covers graded utilities such as execution score or preference reward.

Task utility is usually observed through evaluation signals. A programmatic verifier $V_P(x,d)\in\mathcal{V}$ may return a Boolean decision, a scalar score, a diagnostic object, or a canonicalized answer representation. A learned evaluator $J_\phi(x,\omega)$ scores either a completed leaf $\omega=y$ or a partial state $\omega=z$. Unlike programmatic checks, learned evaluators are proxies whose validity depends on their supervision and test-time protocol~\citep{cobbe2021verifiers,lightman2023verify,zheng2023judge,zhou2025jetts}.

For a task distribution $\mathcal{P}$, the associated test-time scaling curve is
\[
G(B)=\mathbb{E}_{x\sim\mathcal{P},\,\xi}\bigl[U_x(\hat{o}_B(x;\xi))\bigr],
\]
where $\xi$ denotes the algorithm's internal randomness. The term \emph{scaling} refers to how performance varies with budget; it does not imply that performance must be monotone for every method.

\begin{figure}[t]
\centering
\begin{minipage}[t]{0.31\linewidth}
\centering
\begin{tikzpicture}[
  x=0.82cm,
  y=0.82cm,
  >=Latex,
  font=\scriptsize,
  tree/.style={draw=black!28, line width=0.45pt},
  seqedge/.style={draw=green!45!black, line width=1.25pt},
  nod/.style={circle, draw=black!55, fill=white, minimum size=4.2mm, inner sep=0pt},
  snod/.style={circle, draw=green!45!black, fill=green!12, minimum size=4.3mm, inner sep=0pt},
  box/.style={draw=black!70, fill=black!5, rounded corners=2pt, minimum height=5mm, inner sep=2pt},
  ctl/.style={draw=green!45!black, fill=green!8, rounded corners=2pt, minimum height=5mm, inner sep=2pt},
  leaf/.style={draw=black!55, fill=white, rounded corners=2pt, minimum width=8mm, minimum height=5mm, inner sep=1.5pt}
]
\node[align=center, font=\bfseries\small] at (2.0,0.35) {Single-trajectory\\sequential scaling};
\node[box] (root) at (2.0,-0.45) {$x$};
\node[nod] (a1) at (0.8,-1.4) {};
\node[snod] (b1) at (2.0,-1.4) {};
\node[nod] (c1) at (3.2,-1.4) {};
\node[nod] (a2) at (1.2,-2.45) {};
\node[snod] (b2) at (2.0,-2.45) {};
\node[nod] (c2) at (2.8,-2.45) {};
\node[snod] (b3) at (2.0,-3.5) {};
\node[ctl] (ctl) at (3.8,-2.45) {$\nu_t$};
\node[draw=green!45!black, fill=green!8, rounded corners=2pt, minimum width=8mm, minimum height=5mm, inner sep=1.5pt] (outseq) at (2.0,-4.65) {$\hat{y}$};
\node[align=center] at (2.0,-5.35) {adaptive control\\on one path};

\draw[tree] (root) -- (a1);
\draw[tree] (root) -- (b1);
\draw[tree] (root) -- (c1);
\draw[tree] (b1) -- (a2);
\draw[tree] (b1) -- (b2);
\draw[tree] (b1) -- (c2);

\draw[seqedge] (root) -- (b1) -- (b2) -- (b3) -- (outseq);
\draw[->,seqedge] (ctl) -- (b2);
\end{tikzpicture}
\end{minipage}\hfill
\begin{minipage}[t]{0.34\linewidth}
\centering
\begin{tikzpicture}[
  x=0.82cm,
  y=0.82cm,
  >=Latex,
  font=\scriptsize,
  tree/.style={draw=black!30, line width=0.5pt},
  sample/.style={draw=blue!70!black, line width=1.2pt},
  prefixedge/.style={draw=orange!85!black, line width=1.2pt},
  pruned/.style={draw=red!70!black, line width=0.9pt, dashed},
  nod/.style={circle, draw=black!55, fill=white, minimum size=4.4mm, inner sep=0pt},
  box/.style={draw=black!70, fill=black!5, rounded corners=2pt, minimum height=5mm, inner sep=2pt},
  leaf/.style={draw=black!55, fill=white, rounded corners=2pt, minimum width=8mm, minimum height=5mm, inner sep=1.5pt}
]
\node[font=\bfseries\small] at (1.9,0.35) {Leaf-level scaling};
\node[box] (root) at (1.9,-0.45) {$x$};
\node[nod] (a) at (0.7,-1.4) {};
\node[nod] (b) at (1.9,-1.4) {};
\node[nod] (c) at (3.1,-1.4) {};
\node[leaf] (y1) at (0.4,-2.55) {$y_1$};
\node[leaf] (y2) at (1.4,-2.55) {$y_2$};
\node[leaf] (y3) at (2.4,-2.55) {$y_3$};
\node[leaf] (y4) at (3.4,-2.55) {$y_4$};
\node[draw=blue!70!black, fill=blue!8, rounded corners=2pt, minimum height=5mm, inner sep=2pt] (reducer) at (4.6,-3.6) {$\mathcal{R}$};
\node[draw=blue!70!black, fill=blue!8, rounded corners=2pt, minimum width=8mm, minimum height=5mm, inner sep=1.5pt] (outleaf) at (5.8,-3.6) {$\hat{y}$};
\node[align=center] (samples) at (1.9,-3.55) {$N$ complete leaves};

\draw[tree] (root) -- (a);
\draw[tree] (root) -- (b);
\draw[tree] (root) -- (c);
\draw[tree] (a) -- (y1);
\draw[tree] (b) -- (y2);
\draw[tree] (b) -- (y3);
\draw[tree] (c) -- (y4);

\draw[sample] (root) -- (a) -- (y1);
\draw[sample] (root) -- (b) -- (y2);
\draw[sample] (root) -- (b) -- (y3);
\draw[sample] (root) -- (c) -- (y4);
\draw[->,sample] (samples) -- (reducer);
\draw[->,sample] (reducer) -- (outleaf);
\end{tikzpicture}
\end{minipage}\hfill
\begin{minipage}[t]{0.31\linewidth}
\centering
\begin{tikzpicture}[
  x=0.82cm,
  y=0.82cm,
  >=Latex,
  font=\scriptsize,
  tree/.style={draw=black!30, line width=0.5pt},
  sample/.style={draw=blue!70!black, line width=1.2pt},
  prefixedge/.style={draw=orange!85!black, line width=1.2pt},
  pruned/.style={draw=red!70!black, line width=0.9pt, dashed},
  nod/.style={circle, draw=black!55, fill=white, minimum size=4.4mm, inner sep=0pt},
  pnode/.style={circle, draw=orange!85!black, fill=orange!15, minimum size=4.6mm, inner sep=0pt},
  box/.style={draw=black!70, fill=black!5, rounded corners=2pt, minimum height=5mm, inner sep=2pt},
  leaf/.style={draw=black!55, fill=white, rounded corners=2pt, minimum width=8mm, minimum height=5mm, inner sep=1.5pt}
]
\node[font=\bfseries\small] at (1.9,0.35) {Prefix-level scaling};
\node[box] (root) at (1.9,-0.45) {$x$};
\node[pnode] (z1) at (0.7,-1.4) {};
\node[pnode] (z2) at (1.9,-1.4) {};
\node[pnode] (z3) at (3.1,-1.4) {};
\node[pnode] (z21) at (1.4,-2.55) {};
\node[pnode] (z22) at (2.4,-2.55) {};
\node[draw=orange!85!black, fill=orange!12, rounded corners=2pt, minimum width=8mm, minimum height=5mm, inner sep=1.5pt] (outleaf) at (2.4,-3.7) {$\hat{y}$};
\node[align=center] at (1.9,-4.45) {search over prefixes\\before completion};

\draw[tree] (root) -- (z1);
\draw[tree] (root) -- (z2);
\draw[tree] (root) -- (z3);
\draw[tree] (z2) -- (z21);
\draw[tree] (z2) -- (z22);

\draw[prefixedge] (root) -- (z1);
\draw[prefixedge] (root) -- (z2);
\draw[prefixedge] (root) -- (z3);
\draw[prefixedge] (z2) -- (z21);
\draw[prefixedge] (z2) -- (z22);
\draw[prefixedge] (z22) -- (outleaf);

\draw[pruned] (z1) -- ++(-0.35,-0.65);
\draw[pruned] (z3) -- ++(0.35,-0.65);
\draw[pruned] (z21) -- ++(-0.25,-0.7);
\draw[red!70!black, line width=0.9pt] ($(z1)+(-0.12,-0.12)$) -- ($(z1)+(0.12,0.12)$);
\draw[red!70!black, line width=0.9pt] ($(z1)+(-0.12,0.12)$) -- ($(z1)+(0.12,-0.12)$);
\draw[red!70!black, line width=0.9pt] ($(z3)+(-0.12,-0.12)$) -- ($(z3)+(0.12,0.12)$);
\draw[red!70!black, line width=0.9pt] ($(z3)+(-0.12,0.12)$) -- ($(z3)+(0.12,-0.12)$);
\draw[red!70!black, line width=0.9pt] ($(z21)+(-0.12,-0.12)$) -- ($(z21)+(0.12,0.12)$);
\draw[red!70!black, line width=0.9pt] ($(z21)+(-0.12,0.12)$) -- ($(z21)+(0.12,-0.12)$);
\end{tikzpicture}
\end{minipage}
\caption{Three regimes of test-time scaling over the implicit prefix tree. Left: single-trajectory sequential scaling allocates compute to one evolving response without maintaining a competing frontier. Middle: leaf-level scaling allocates compute to completed root-to-leaf trajectories and reduces them to one output. Right: prefix-level scaling allocates compute to internal prefixes, expanding promising continuations and pruning others before committing to a completed leaf.}
\label{fig:leaf-vs-prefix-scaling}
\end{figure}
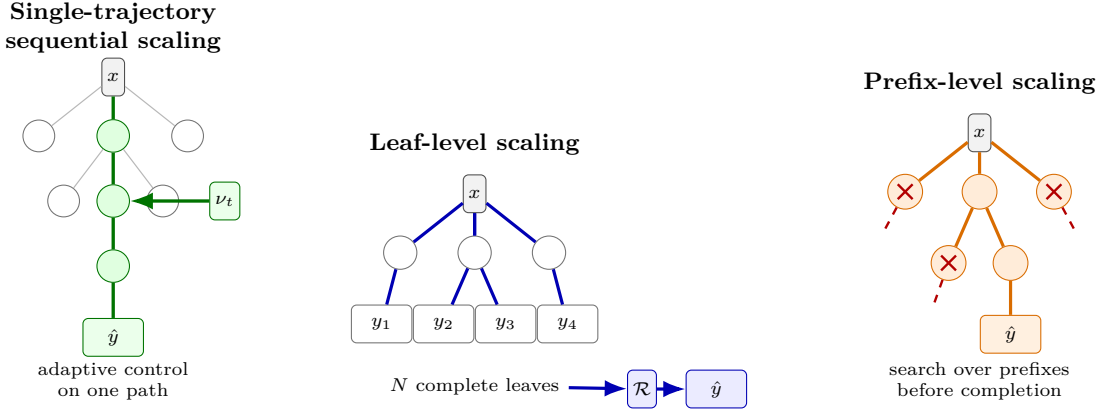

We use the following operational taxonomy, illustrated in \Cref{fig:leaf-vs-prefix-scaling}.

\paragraph{Single-trajectory sequential scaling.}
At every step, there is at most one unfinished active prefix. Additional computation changes how that prefix is extended, revised, or terminated, but never branches into a competing frontier.

\paragraph{Leaf-level scaling.}
Additional compute produces a bank of completed candidates. Any interaction between candidate trajectories is deferred until after the candidates have completed, via a terminal reducer.

\paragraph{Prefix-level scaling.}
Additional compute is allocated based on scores for unfinished prefixes. Expansion, pruning, rollout allocation, or termination decisions depend on partial states before completion.

These regimes are not mutually exclusive at the system level: many practical methods are hybrids, most commonly prefix search followed by a leaf-level reducer.

\subsection{Single-trajectory sequential scaling}
\label{sec:form:singletraj}

A single-trajectory method maintains one active state $z_t$ and remaining budget $b_t$. At step $t$, a controller chooses a meta-action
\[
\nu_t \sim \pi_{\mathrm{seq}}(\cdot \mid x,z_t,b_t),
\]
then extends only the active path,
\[
z_{t+1} \sim \mathrm{Extend}(z_t;\nu_t,p_\theta),
\qquad
b_{t+1} = b_t - c(\nu_t).
\]
Actions include continuing generation, suppressing \textsc{eos}, appending a control string or critique, changing decoding hyperparameters, or stopping. Budget-forcing methods such as \textsc{s1} fit this template. Inference-time meta-generation procedures also fit when they maintain one evolving response rather than a branching frontier~\citep{muennighoff2025s1,welleck2024inference_time_algos}.

This regime has low orchestration overhead because all extra compute goes to one candidate. Without a competing unfinished prefix, however, recovery from an early misconception requires self-revision along the same path rather than a search over alternatives.

\subsection{Leaf-level scaling: sampling and reduction}
\label{sec:form:leaflevel}

In leaf-level scaling, additional compute produces completed leaves and a terminal reducer chooses the output. Write $[N]=\{1,\ldots,N\}$. In the canonical case,
\[
\mathcal{Y}_N(x)=\{Y_i\}_{i=1}^N,
\qquad
Y_i \overset{\mathrm{i.i.d.}}{\sim} q_\pi(\cdot\mid x),
\]
where $q_\pi$ is a common proposal distribution induced by the base model and decoding policy. Appendix~\ref{app:leaf_reducer_variants} extends this template to weighted and heterogeneous-proposal banks. For each candidate, let $m_i$ contain signals available at inference time, and define
\[
(r_i,a_i)=\mathrm{Parse}(Y_i),
\qquad
(d_i,s_i)=\Psi(Y_i),
\qquad
v_i = V_P(x,d_i),
\qquad
j_i = J_\phi(x,Y_i).
\]
Given the proposal, the terminal stage is specified by a reducer
\[
\mathcal{R}_N:
\bigl(x,\{(Y_i,m_i,v_i,j_i)\}_{i=1}^N\bigr)
\mapsto
\hat{y}\ \text{or}\ \hat{a}.
\]
The fixed-bank leaf template separates generation from reduction. Scores assigned to one unfinished candidate cannot affect another; cross-candidate interaction begins inside $\mathcal{R}_N$ after all $N$ leaves are complete. Adaptive stopping preserves this separation between unfinished candidates. A controller may use completed leaves to decide whether to launch another independent rollout, or it may truncate a rollout using only that rollout's prefix-local evidence. It does not reallocate compute among competing unfinished prefixes (\Cref{ssec:aggregation}). Such controllers change the realized sample count or trace length and are not fixed-bank reducers.

\subsubsection{Canonical reduction rules}

\paragraph{Verifier-constrained selection.}
Let $\mathcal{I}_{\mathrm{pass}}=\{i:V_P(x,d_i)\in \mathrm{PASS}\}$. A verifier-constrained reducer selects
\[
i^\star \in \arg\max_{i\in\mathcal{I}_{\mathrm{pass}}} T(x,Y_i,v_i,j_i),
\]
where $T$ ranks the passing candidates using, for example, the learned score $j_i$ or sequence likelihood. The reducer uses a declared fallback if $\mathcal{I}_{\mathrm{pass}}$ is empty. This is the generate-test-select pattern used in program-synthesis and verifier-based reasoning systems~\citep{li2022alphacode,cobbe2021verifiers}.

For answer-valued tasks, define the eligible set
\(\mathcal{I}_{\mathrm{valid}}=\{i\in[N]:a_i\ne\bot\}\). For tasks without an
invalid-answer state, take \(\mathcal{I}_{\mathrm{valid}}=[N]\). Every reducer
must declare a deterministic tie rule and a fallback output for
\(\mathcal{I}_{\mathrm{valid}}=\emptyset\).

\paragraph{Answer marginalization and self-consistency.}
When answers or verifier outputs can be canonicalized and
\(\mathcal{I}_{\mathrm{valid}}\ne\emptyset\), define the empirical answer
distribution over eligible candidates
\[
\hat{q}_N(a\mid x)=\frac{1}{|\mathcal{I}_{\mathrm{valid}}|}
\sum_{i\in\mathcal{I}_{\mathrm{valid}}}\mathbf{1}[a_i=a],
\qquad a\ne\bot.
\]
Self-consistency selects
\[
\hat{a} \in \arg\max_{a\ne\bot} \hat{q}_N(a\mid x).
\]
Appendix~\ref{app:leaf_reducer_variants} shows that this is exactly empirical MBR with agreement utility $u(a,a')=\mathbf{1}[a=a']$~\citep{wang2022selfconsistency,kumar2004mbr,bertsch2023mbr}.

\paragraph{Learned reranking (Best-of-$N$).}
When programmatic verification is unavailable or incomplete, a standard reducer chooses among eligible candidates
\[
i^\star \in \arg\max_{i\in\mathcal{I}_{\mathrm{valid}}} j_i.
\]
This covers learned verifiers, reward models, and LLM judges used as
inference-time rerankers~\citep{cobbe2021verifiers,snell2025computeoptimal,
huang2025bestofn,zhou2025jetts}.

\paragraph{Empirical expected-utility selection.}
More generally, with a projection $\psi$ and utility $u$, one may select
\[
i^\star \in \arg\max_{i\in\mathcal{I}_{\mathrm{valid}}}
\frac{1}{|\mathcal{I}_{\mathrm{valid}}|}
\sum_{k\in\mathcal{I}_{\mathrm{valid}}}
u\!\left(\psi(Y_i),\psi(Y_k)\right).
\]
This MBR template unifies plurality vote, semantic-consensus rules, and
neural-metric reranking under a common expected-utility
objective~\citep{kumar2004mbr,freitag2022neuralmbr,bertsch2023mbr}.

These rules cover the main structural forms of leaf-level reduction. Appendix~\ref{app:leaf_reducer_variants} details weighted aggregation, semantic kernels, pairwise aggregation, regularized selection, and the separation of answer selection from rationale presentation.

\subsubsection{Failure modes and budget allocation}

Increasing the number of samples from a fixed proposal expands candidate coverage without early pruning, but the reducer introduces its own failure modes. An incomplete verifier can admit false positives as $N$ grows, while hard argmax selection can overoptimize a misspecified learned score. This produces the non-monotone Best-of-$N$ behavior documented in work on reward-model overoptimization and inference-time reward hacking~\citep{gao2023scaling,huang2025bestofn,khalaf2025rewardhacking}. Judge-based reducers also inherit position and verbosity biases, so judge design and prompting are part of the method under evaluation~\citep{zheng2023judge,zhou2025jetts}. Leaf count and evaluator-call count are separate budget axes: MBR-style reducers may require $O(N^2)$ utility calls, which motivates approximations such as confidence-based pruning and approximate MBR~\citep{cheng2023cbp,jinnai2024ambr}.

A fixed sample count is therefore only one leaf-level design choice. Let
$\hat{o}_n(x)$ denote the output of the declared size-$n$ leaf algorithm. One
may instead choose $N=N(x)$ by solving
\[
N(x) \in \arg\max_{n:\, C_{\mathrm{gen}}(n)+C_{\mathrm{eval}}(n)\le B}
\mathbb{E}\bigl[U_x(\hat{o}_n(x))\bigr],
\]
where $C_{\mathrm{gen}}$ denotes generation cost and $C_{\mathrm{eval}}$ covers
inference-time scoring, control, and reduction. Recent studies of inference-time scaling
optimize budget allocation across strategies~\citep{wu2024inference};
difficulty-adaptive allocation can outperform fixed-$N$
baselines~\citep{snell2025computeoptimal}.

\subsection{Prefix-level scaling: search over partial states}
\label{sec:form:prefixlevel}

Prefix-level scaling allocates compute before trajectories are complete. To cover both token-level search and step-level methods such as Tree-of-Thoughts, RAP, and AlphaZero-like decoding, let $z$ denote a search state representing either a literal token prefix or a macro-prefix composed of one or more reasoning steps, and write $\mathrm{Succ}(z)$ for its allowed expansions~\citep{yao2023tot,hao2023rap,wan2024alphazero}.

\subsubsection{Frontier, terminal bank, and budget}

A prefix-level method maintains an active frontier $\mathcal{F}_t$, a bank of completed leaves $\mathcal{Y}_t$, and remaining budget $b_t$. A generic iteration is
\[
z_t \in \mathrm{Select}(\mathcal{F}_t)
\;\to\;
\mathcal{C}_t \subseteq \mathrm{Succ}(z_t)
\;\to\;
\text{score } \mathcal{C}_t
\;\to\;
(\mathcal{F}_{t+1},\mathcal{Y}_{t+1},b_{t+1}).
\]
The budget may count generated tokens, node expansions, rollout calls, verifier or judge calls, or wall-clock cost. Unlike leaf-level scaling, prefix-level search uses evaluations of unfinished states to allocate compute among competing prefixes.

\subsubsection{Prefix evaluation as continuation-value estimation}

When a search commits to a single prefix and returns one completion drawn from it, the Bayes-optimal prefix score is the continuation value
\[
Q^\star_\rho(z)=\mathbb{E}_{Y\sim q_\rho(\cdot\mid x,z)}\bigl[U_x(Y)\bigr],
\]
where $q_\rho(\cdot\mid x,z)$ denotes the distribution over completed leaves obtained by continuing from prefix $z$ with rollout policy $\rho$. If search instead accumulates a bank that is later passed to a reducer $\mathcal{R}$, the exact marginal value of expanding $z$ becomes history-dependent because it depends on how future leaves from $z$ will interact with the leaves already collected. One-leaf surrogates such as correctness probability, process reward, or rollout return ignore this coupling.

\subsubsection{Canonical prefix evaluators}

\paragraph{Likelihood-based scoring.}
A canonical heuristic is cumulative log-probability under the local generation policy:
\[
S_{\mathrm{LL}}(z)=\sum_{j=1}^{|z|} \log \pi(z_j\mid x,z_{<j}).
\]
Implementations sometimes modify this score with a length penalty. With $\pi=p_\theta$, the unnormalized form is the standard beam-search score and ranks prefixes by proposal probability. Length-penalized variants need not preserve that ranking. Neither form directly measures downstream utility.

\paragraph{Outcome-supervised value models.}
Outcome-supervised value models approximate the probability that a prefix reaches a correct completion and use that estimate for pruning, either directly or in combination with likelihood:
\[
\widehat{Q}(z) \approx \Pr_{Y\sim q_\rho(\cdot\mid x,z)}\!\bigl(U_x(Y)=1\bigr),
\qquad
S(z)=\widehat{Q}(z)
\ \text{or}\ 
S(z)=\alpha S_{\mathrm{LL}}(z)+\beta\widehat{Q}(z).
\]
Such value-guided decoding is explored for mathematical reasoning in OVM-style systems~\citep{yu2024ovm}.

\paragraph{Process reward models (PRM).}
When supervision is available on intermediate reasoning steps, a process reward model can score a partial path directly. If $z=(s_1,\ldots,s_t)$ is a sequence of steps, one may use
\[
S_{\mathrm{PRM}}(z)=\sum_{k=1}^t R_\phi(x,s_{\le k})
\qquad \text{or} \qquad
S_{\mathrm{PRM}}(z)=R_\phi(x,z),
\]
where $R_\phi$ scores each step in the context of the problem and the preceding steps and is learned from step-level supervision~\citep{lightman2023verify}.

\paragraph{Rollout-based evaluation.}
When learned value estimates are unreliable, continuation value can be estimated by Monte Carlo completion:
\[
\widehat{Q}_{\mathrm{MC}}(z)=\frac{1}{M_{\mathrm{roll}}}\sum_{m=1}^{M_{\mathrm{roll}}} U_x\!\left(Y^{(m)}\right),
\qquad
Y^{(m)}\sim q_\rho(\cdot\mid x,z).
\]
MCTS-style reasoning methods can be interpreted as structured rollout allocation together with backup of these estimates~\citep{hao2023rap,wan2024alphazero}.

\paragraph{Sound partial checks.}
A partial check $H(x,z)\in\{0,1\}$ is \emph{sound for rejection} if
\[
H(x,z)=0 \;\Longrightarrow\; U_x(y)=0 \quad \text{for all } y \succeq z.
\]
Only sound prefix tests justify safe pruning. Learned value or reward models generally do not provide this guarantee.

\subsubsection{Score-induced search procedures}

\paragraph{Beam search.}
Beam search keeps a frontier of size $K$ and updates it by
\[
\mathcal{F}_{t+1} = \mathrm{TopK}\Bigl(\bigcup_{z\in\mathcal{F}_t} \mathrm{Succ}(z)\,;\,S(\cdot)\Bigr).
\]
Standard beam search uses $S=S_{\mathrm{LL}}$, while verifier-guided and
value-guided variants replace or augment likelihood with learned prefix
scores~\citep{yu2024ovm}.

\paragraph{Best-first search.}
Best-first search repeatedly expands the highest-scoring active prefix,
\[
z^\star \in \arg\max_{z\in\mathcal{F}_t} S(z),
\qquad
\mathcal{F}_{t+1} = (\mathcal{F}_t\setminus\{z^\star\}) \cup \mathrm{Succ}(z^\star).
\]
In classical search, admissible heuristics can yield optimality guarantees. In
LLM-agent search, learned model-based value functions can be used
heuristically; Koh et al.\ provide a best-first
example~\citep{koh2025treesearchagents}.

\paragraph{MCTS-style search.}
Monte Carlo tree search alternates selection, expansion, evaluation, and backup. A standard UCT rule selects
\[
a_t \in \arg\max_a \left(
Q(z,a) + c_{\mathrm{uct}}\sqrt{\frac{\log n(z)}{n(z,a)}}
\right),
\]
where $n(z)$ and $n(z,a)$ are visit counts and unvisited actions are selected first. Policy-prior variants such as PUCT replace the pure UCT bonus with one modulated by an action prior. In LLM reasoning, RAP, AlphaZero-like search, and related methods instantiate this template with rollout-based or learned value estimates~\citep{kocsis2006uct,browne2012mcts,hao2023rap,wan2024alphazero}. Tree-of-Thoughts can be viewed as a related search controller operating over thought-level rather than token-level expansions~\citep{yao2023tot}.

Additional controller variants, including learned search policies and reward-guided objectives, are discussed in Appendix~\ref{app:prefix_search_variants}.

\subsubsection{Scaling behavior and comparison to leaf-level methods}

Prefix-level search can use compute more efficiently than unguided leaf sampling when prefix scores concentrate budget on high-value branches. Its risk is that pruning changes the support of reachable completions. If every ancestor of a correct leaf is pruned, no later computation can recover that leaf. For pruning, the within-instance ordering of competing frontier states often matters more than global calibration.

Under a fixed budget $B$, prefix search induces an effective proposal distribution over completed leaves,
\[
q_{\mathrm{search},B}(y\mid x)=\Pr\bigl(\hat{y}_B(x)=y\bigr),
\]
which depends jointly on the scorer, controller, and stopping rule. When search
returns an ordered bank or multiset $\mathcal{Y}$ rather than one leaf, its
corresponding object is the joint law
$Q_{\mathrm{search},B}(\mathcal{Y}\mid x)$; per-leaf inclusion probabilities
are only marginals and do not determine bank size, multiplicity, order, or
dependence. Candidate banks returned by prefix search are therefore not i.i.d.\
draws from $q_\pi(\cdot\mid x)$, so direct comparisons to leaf-level
Best-of-$N$ must account for generation, evaluation, control, and decision costs.

Prefix search creates branching structures with shared prefixes. Tree-structured inference kernels can reuse key-value caches across these prefixes and thereby change the compute frontier; DEFT is one example~\citep{yao2025deft}. A hybrid system can first construct a search-induced bank of completed leaves and then apply a leaf-level reducer. Such banks can be summarized by the repeated-sampling metrics in \Cref{ssec:tts_metrics}.

\section{Evaluating Test-Time Scaling}
\label{sec:eval}

Under test-time scaling, reported performance depends on the \emph{full inference protocol} defined over the prefix tree in \Cref{sec:form:setup_taxonomy}. The protocol determines both the \emph{candidate distribution} and the final output rule. Its specification includes budget allocation, decoding or search, inference-time evidence, aggregation and stopping rules, and any verifier or judge calls. A report should state the benchmark structure, the induced \emph{proposal distribution}, the aggregation and stopping protocol, the reported \emph{estimand}, and the uncertainty procedure.

\subsection{Benchmark design as evaluation structure}

Benchmark design determines which capability is measured and whether additional inference-time compute can change outcomes. Under test-time scaling, benchmark suites should be organized by \emph{evaluation structure} as well as task domain.

One distinction is between \emph{verifiable} and \emph{open-ended} tasks. In verifiable settings such as mathematics and code generation, correctness can be checked exactly or by execution, so repeated sampling and terminal reduction target a directly checkable criterion~\citep{chen2021evaluating,li2022alphacode}. In open-ended settings, evaluation relies on rubrics, pairwise preferences, or judge models. Prompts, comparison order, side randomization, and aggregation can change the scores and must be reported with the results~\citep{liang2023helm,zheng2023judge,chiang2024chatbot}. Additional compute also affects reducers differently: verifier-based selection, agreement-based aggregation, and judge-based reranking need not improve at the same rate as the budget increases~\citep{cobbe2021verifiers,lightman2023verify,zheng2023judge}.

The benchmark structure should also match the task's interaction model. Single-turn exact-answer tasks probe discovery of a correct terminal leaf, whereas interactive agent benchmarks evaluate multi-step user/tool interactions and their terminal state~\citep{yao2024taubench}. Inference-time algorithms may also use environment feedback, backtracking, or iterative refinement during a trajectory~\citep{welleck2024inference_time_algos}. The same inference algorithm may score well under ``any correct sample'' reporting but poorly under trajectory-level reliability metrics (\Cref{ssec:tts_metrics}).

Benchmark difficulty should leave room for a scaling curve: if tasks are too easy, additional compute has little effect; if they are uniformly difficult, correct solutions are rarely reached and the curve remains flat~\citep{snell2025computeoptimal}. Because many reasoning benchmarks are small, dataset size limits the statistical resolution of scaling comparisons~\citep{miller2024adding,ye2024benchmarking} (see \Cref{subsec:eval_reproducibility}).

\subsection{Decoding protocols and induced proposal distributions}
\label{ssec:sampling}

Under test-time scaling, the inference system samples from the proposal distribution induced by a complete
decoding protocol rather than directly from the raw model distribution $p_\theta$. Let $\Sigma$ denote
the token vocabulary. At a partial generation
$h_t=(x,y_{<t})$, the model produces logits $\ell_t\in\mathbb{R}^{|\Sigma|}$ and token probabilities
\[
p_t(i)=\frac{\exp \ell_t(i)}{\sum_{j\in\Sigma}\exp \ell_t(j)},\qquad i\in\Sigma.
\]
A one-token decoder with configuration $\lambda$ and finite state $\sigma_t$ returns
\[
(\widetilde q_t,a^{\mathrm{tok}}_t,\sigma_{t+1})
  =D_\lambda(h_t,\ell_t,p_t,\chi_t,\sigma_t),
\]
where $\chi_t$ denotes side information available to the decoder, such as hidden states, token embeddings,
auxiliary scores, rollout estimates, or controller state. If $a^{\mathrm{tok}}_t\neq \bot$, the next token is set
deterministically to $y_t=a^{\mathrm{tok}}_t$; otherwise $y_t\sim \operatorname{Cat}(\widetilde q_t)$.

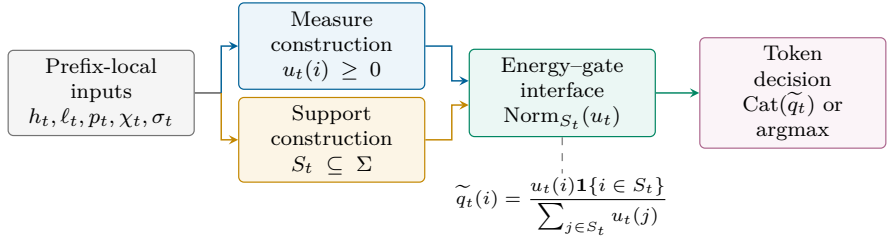
\begin{figure}[t]
\centering
\begin{tikzpicture}[
  font=\footnotesize,
  >=stealth,
  box/.style={
    rounded corners=2pt,
    align=center,
    inner sep=3pt,
    minimum height=1.05cm,
    text width=2.25cm,
    line width=0.45pt
  },
  inputbox/.style={box, draw=black!55, fill=black!4},
  measurebox/.style={box, draw=oiBlue!85!black, fill=oiBlue!7},
  supportbox/.style={box, draw=oiOrange!85!black, fill=oiOrange!10},
  gatebox/.style={box, draw=oiGreen!85!black, fill=oiGreen!8},
  decisionbox/.style={box, draw=oiPurple!85!black, fill=oiPurple!8},
  formula/.style={align=center, font=\scriptsize, inner sep=1pt},
  flow/.style={->, line width=0.55pt, draw=black!65},
  link/.style={line width=0.55pt, draw=black!65}
]
\node[inputbox] (inputs) at (0,0) {Prefix-local\\ inputs\\ $h_t,\ell_t,p_t,\chi_t,\sigma_t$};
\node[measurebox] (measure) at (3.05,0.62) {Measure\\ construction\\ $u_t(i)\ge 0$};
\node[supportbox] (support) at (3.05,-0.62) {Support\\ construction\\ $S_t\subseteq\Sigma$};
\node[gatebox] (gate) at (6.10,0) {Energy--gate\\ interface\\ $\operatorname{Norm}_{S_t}(u_t)$};
\node[decisionbox] (decision) at (9.15,0) {Token\\ decision\\ $\operatorname{Cat}(\widetilde q_t)$ or argmax};

\coordinate (split) at ($(inputs.east)+(0.35,0)$);
\draw[link] (inputs.east) -- (split);
\draw[flow, draw=oiBlue!85!black] (split) |- (measure.west);
\draw[flow, draw=oiOrange!85!black] (split) |- (support.west);
\draw[flow, draw=oiBlue!85!black] (measure.east) -- ++(0.35,0) |- ([yshift=0.16cm]gate.west);
\draw[flow, draw=oiOrange!85!black] (support.east) -- ++(0.35,0) |- ([yshift=-0.16cm]gate.west);
\draw[flow, draw=oiGreen!85!black] (gate) -- (decision);

\node[formula, below=0.48cm of gate] (eq) {$\displaystyle
\widetilde q_t(i)=
\frac{u_t(i)\mathbf{1}\{i\in S_t\}}
{\sum_{j\in S_t}u_t(j)}
$};
\draw[link, dashed, draw=black!45] (gate.south) -- (eq.north);
\end{tikzpicture}
\caption{Elementwise view of a one-token sampling protocol. The sequence-level proposal distribution
$q_\pi(\cdot\mid x)$ is induced by composing these local kernels until termination.}
\label{fig:energy-gate-decoder}
\end{figure}

A broad class of stochastic decoders admits an energy--gate form (\Cref{fig:energy-gate-decoder}). For a nonnegative token measure
$u_t:\Sigma\to\mathbb{R}_{\ge 0}$ and support $S_t\subseteq \Sigma$ with
$\sum_{j\in S_t}u_t(j)>0$,
\begin{equation}
\label{eq:energy-gate-normalization}
\widetilde q_t(i)
=
\operatorname{Norm}_{S_t}(u_t)_i
=
\frac{u_t(i)\mathbf{1}\{i\in S_t\}}
{\sum_{j\in S_t}u_t(j)}.
\end{equation}
The measure $u_t$ determines relative weights, while the support $S_t$ determines which tokens remain
eligible. Ordinary temperature sampling uses $u_t(i)=\exp(\ell_t(i)/T)$ and $S_t=\Sigma$; ordinary
truncation rules restrict $S_t$ and renormalize the same measure; uniform or coverage-based samplers
change the weights inside the retained set. Thus support equivalence and distribution equivalence are
distinct: two decoders can retain the same candidate set while inducing different token probabilities.
If a gate retains zero measure, the decoder must invoke a declared fallback
before applying \(\operatorname{Norm}\); otherwise its token kernel is undefined.

\FloatBarrier

We treat EOS or STOP as an absorbing action and require
termination almost surely; a hard length cap or causal truncation maps the
current prefix to an explicit terminal outcome. Under these conditions,
composing the local kernels gives the leaf-level proposal distribution used in
\Cref{sec:form:leaflevel}:
\begin{equation}
\label{eq:sequence-proposal}
q_\lambda(y\mid x)
=
\prod_{t=1}^{\tau(y)}
K_{\lambda,t}(y_t\mid x,y_{<t},\sigma_t),
\end{equation}
where $\tau(y)$ denotes the termination step of $y$ (its length in tokens) and $K_{\lambda,t}$ is a point
mass when the decoder takes a deterministic token action and otherwise equals $\widetilde q_t$. When the
protocol $\pi$ fixes a single configuration $\lambda$, we write $q_\pi=q_\lambda$; if the protocol randomizes
over configurations, $q_\pi$ averages $q_\lambda$ over that outer randomness; if a bank uses heterogeneous
protocols, then candidate $i$ is drawn from its own proposal $q_{\pi_i}(\cdot\mid x)$ rather than from a
common i.i.d.\ proposal.

Support construction captures the common part of many truncation rules. Each gate below takes the live
candidate set $A$ and a reference token score $r$, typically $p_t$ or a monotone transform of it; under
sequential composition, the renormalized $r^{(j)}$ in
\Cref{eq:sequential-gates} plays this role. We use three gate families. A
\emph{level gate} keeps tokens above a data-dependent threshold,
\begin{equation}
\label{eq:level-gate}
G_{\mathrm{lev}}(A;r,g,\tau_g)
=
\{i\in A: g(i;r,\ell_t,\chi_t)\ge \tau_g(r,\ell_t,\chi_t)\}.
\end{equation}
This family includes absolute-probability, mode-relative, entropy-scaled, and logit-band thresholds. A
\emph{head-budget gate} sorts the live set $A$ in increasing order of a cost key $\kappa$ (for
probability-ranked truncation, $\kappa(i)=-r(i)$) and retains the first $m_\beta$ tokens permitted by the
gate parameter $\beta$ (a fixed rank for top-$k$; the smallest count whose retained mass reaches the target
for cumulative-mass rules),
\begin{equation}
\label{eq:head-gate}
G_{\mathrm{head}}(A;r,\kappa,\beta)
=
\{i_{(1)},\ldots,i_{(m_\beta)}\},
\qquad
\kappa(i_{(1)})\le \cdots \le \kappa(i_{(|A|)}).
\end{equation}
This family includes fixed-rank, cumulative-mass, curvature, entropy-budget, and typicality-based truncation. An
\emph{objective gate} chooses a support by solving, or approximating, a set objective,
\begin{equation}
\label{eq:objective-gate}
G_{\mathrm{obj}}(A;r,J)
\in
\arg\min_{\emptyset\neq A'\subseteq A} J(A';r,\chi_t).
\end{equation}
This separates the support decision from the weighting rule in
\Cref{eq:energy-gate-normalization}.

A special case is the family of mode-preserving head gates. Fix a reference distribution $r_t$ whose
rank order is the model rank order, for example $p_t$ or a positive-temperature rescaling, and write
\[
r_t(i_1)\ge r_t(i_2)\ge\cdots\ge r_t(i_{|\Sigma|}),
\qquad
K_m=\{i_1,\ldots,i_m\}.
\]
If each gate $g$ in a parallel composition $\mathcal H$ returns a model-ranked prefix $K_{m_g}$, then the
composed support collapses to the strictest retained prefix:
\begin{equation}
\label{eq:head-intersection}
\bigcap_{g\in\mathcal H} K_{m_g}
=
K_{m_\cap},
\qquad
m_\cap=\min_{g\in\mathcal H} m_g.
\end{equation}
Thus, any parallel combination of top-$k$~\citep{fan2018hierarchical},
nucleus~\citep{holtzman2020curious}, probability-threshold~\citep{hewitt2022truncation}, mode-relative, or
logit-band head gates has the common form
\begin{equation}
\label{eq:parallel-head-distribution}
\widetilde q_t^{\mathcal H}(i)
=
\frac{u_t(i)\mathbf{1}\{i\in K_{m_\cap}\}}
{\sum_{j=1}^{m_\cap}u_t(i_j)}.
\end{equation}
The different named rules specify different ways of computing $m_g$; once their head supports are evaluated
in parallel, only the smallest retained prefix remains active.

This collapse does not apply to all decoders. Locally typical sampling can skip high-probability tokens
because it orders tokens by the distance of their surprisal $-\log p_t(i)$ from the conditional entropy of
$p_t$ rather than by model probability~\citep{meister2023typical}; geometry- or
objective-based crops can select non-prefix supports; and measure-changing rules can reorder candidates even
when the support is fixed. Ordered composition also differs from parallel masking. For probability-only gates,
a sequential protocol has the form
\begin{equation}
\label{eq:sequential-gates}
A_0=\Sigma,\qquad
r^{(0)}=p_t,\qquad
A_j=G_j(A_{j-1};r^{(j-1)},\ell_t,\chi_t),\qquad
r^{(j)}=\operatorname{Norm}_{A_j}(w^{(j)}),
\end{equation}
where $w^{(j)}$ is the measure exposed by the $j$-th operator. Thus top-$p$ followed by top-$k$ need not
match top-$k$ followed by top-$p$, because cumulative masses, entropies, thresholds, and ranks may be
recomputed after renormalization. Temperature before a mass gate can change the support, whereas
temperature after a fixed support only changes within-support weights. Each
sequential gate must leave positive exposed mass, or apply its declared fallback,
before the next normalization.

For evaluation, these distinctions determine the estimand. In leaf-level scaling with a fixed stochastic
protocol, the usual candidate bank is
\[
Y_1,\ldots,Y_N \stackrel{\mathrm{i.i.d.}}{\sim} q_\pi(\cdot\mid x).
\]
A deterministic configuration sweep instead returns a reproducible set of design points, not Monte Carlo
samples from a single proposal. Prefix-level search induces its own budget-dependent proposal
$q_{\mathrm{search},B}(\cdot\mid x)$ through the scorer, controller, branching rule, rollout policy, and stopping
criterion, and its returned leaves are generally dependent. Accordingly, the repeated-sampling diagnostics of
\Cref{ssec:tts_metrics} have an i.i.d.\ interpretation only when the bank is generated by independent draws from a
declared fixed proposal; otherwise they are descriptive summaries of the returned bank.

A sampling report should specify the token measure, support gates, reference score space, gate
ordering, deterministic actions or stochastic draws, state updates, stopping and length rules, number of
samples, seeds or randomized configuration policy, and any approximation or fallback behavior. These details
define the proposal distribution over reasoning traces and the test-time scaling system being evaluated.

\subsection{Aggregation and stopping under sampled inference}
\label{ssec:aggregation}

\begin{figure}[t]
\centering
\begin{tikzpicture}[
  font=\scriptsize,
  >=Latex,
  box/.style={
    rounded corners=2pt,
    align=center,
    inner sep=3pt,
    minimum height=1.08cm,
    line width=0.45pt
  },
  bank/.style={box, draw=black!60, fill=black!4, text width=2.05cm},
  evidence/.style={box, draw=oiBlue!85!black, fill=oiBlue!7, text width=2.65cm},
  decision/.style={box, draw=oiPurple!85!black, fill=oiPurple!8, text width=2.70cm},
  output/.style={box, draw=oiGreen!85!black, fill=oiGreen!8, text width=1.45cm},
  stop/.style={box, draw=oiOrange!85!black, fill=oiOrange!9, text width=3.55cm, minimum height=0.82cm},
  flow/.style={->, line width=0.55pt, draw=black!65},
  control/.style={->, dashed, line width=0.55pt, draw=oiOrange!90!black}
]
\node[bank] (bank) at (0,0) {Sample stream\\completed leaves $Y_{1:n}$};
\node[evidence] (evidence) at (3.10,0) {Evidence map $E_\eta$\\token/step confidence; verifier, PRM, or judge};
\node[decision] (decision) at (6.65,0) {Fixed-bank decision $\mathcal{G}_\delta$\\vote; max; weighted/gated pooling; MBR};
\node[output] (output) at (9.55,0) {Submitted\\output $\hat{o}$};

\draw[flow] (bank) -- (evidence);
\draw[flow] (evidence) -- (decision);
\draw[flow] (decision) -- (output);

\node[stop] (rollstop) at (2.15,1.55) {Across-rollout controller $\tau_{\mathrm{roll}}$\\reads completed history; saves future rollouts};
\node[stop] (tokstop) at (2.15,-1.55) {Within-rollout controller $\tau_{\mathrm{tok}}$\\reads the current token prefix; saves suffix tokens};
\draw[control] (bank.north) -- ++(0,0.28) -| (rollstop.south);
\draw[control] (rollstop.west) -| ([xshift=-0.25cm]bank.west) -- (bank.west);
\draw[control] (bank.south) -- ++(0,-0.28) -| (tokstop.north);
\draw[control] (tokstop.west) -| ([xshift=-0.25cm]bank.west) -- (bank.west);
\end{tikzpicture}
\caption{Aggregation and stopping under sampled inference. Evidence extraction and fixed-bank selection operate on completed traces, so they cannot save generation already spent. Causal controllers inspect only revealed history: across-rollout stopping avoids later traces, whereas token-prefix stopping can also truncate the current trace. The evidence map, controller, and final decision are all part of the evaluated protocol.}
\label{fig:aggregation-protocol}
\end{figure}
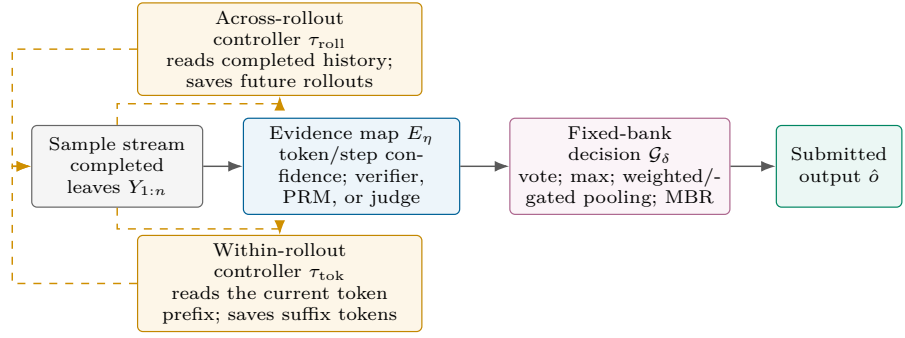

A sampled candidate bank does not by itself define the submitted prediction. The inference protocol must say how evidence is extracted from each trace and how a fixed, completed bank is converted into an output. If either the number or length of rollouts is adaptive, the protocol must also specify when generation stops. Changing the evidence map or decision rule can improve accuracy without reducing generation cost; \Cref{fig:aggregation-protocol} separates this completed-bank pathway from causal controllers that can avoid later rollouts or truncate the current one.

\paragraph{Evidence and fixed-bank decisions.}
For completed candidate $i$, let $m_i$ contain signals available at inference time; the benchmark utility $U_x$ is excluded. Examples include chosen-token log-probabilities, next-token distributions, process-reward scores, and verifier or judge outputs. An evidence map produces an evidence object
\[
e_i=E_\eta(x,Y_i,m_i)\in\mathcal{E},
\]
where $\mathcal{E}$ may be scalar or structured and $E_\eta$ includes any reduction over tokens or reasoning steps required by the downstream rule. We use scalar $e_i\in\mathbb{R}$ for the score-based rules below. Sequence log-likelihood and mean token log-likelihood~\citep{wang2022selfconsistency}, statistics of next-token distributions~\citep{kang2025selfcertainty,fu2026deepconf}, reductions of per-step process scores~\citep{lightman2023verify}, and external verifier or reward-model scores~\citep{cobbe2021verifiers} are distinct evidence maps even when passed to the same selector. Pairwise methods may instead construct
\[
[\mathbf{W}_n]_{ij}
=
 u\!\left(\psi(Y_i),\psi(Y_j)\right).
\]
A fixed-bank decision rule then returns a leaf, an answer, or a set of outputs,
\[
\hat{o}_n
=
\mathcal{G}_\delta\!\left(
 x,\bigl((Y_i,a_i,e_i)\bigr)_{i=1}^{n},\mathbf{W}_n
\right).
\]
Together, $E_\eta$ and $\mathcal{G}_\delta$ refine the generic leaf-level reducer $\mathcal{R}_N$ of \Cref{sec:form:leaflevel} into evidence and decision stages. The evidence source and decision rule are separate choices. Scores from one source can be used for reranking, filtering, or vote weighting. Conversely, a voting rule may take model likelihoods as evidence or replace them with self-evaluations or an external verifier.

For answer-valued tasks, write $a_i=\bot$ when parsing or canonicalization
declares a candidate invalid, let
$\mathcal{I}_{\mathrm{valid}}=\{i:a_i\ne\bot\}$, and let
$I_a=\{i\in\mathcal{I}_{\mathrm{valid}}:a_i=a\}$. Plurality
self-consistency\footnote{We use \emph{plurality voting} because this reducer
selects the most frequent extracted answer even when it receives at most half
of the votes. A strict majority requires more than half, so \emph{majority
voting} can be ambiguous for this rule.} scores a group by
$|I_a|$~\citep{wang2022selfconsistency}. Hard
Best-of-$N$ chooses the largest-$e_i$ candidate in
$\mathcal{I}_{\mathrm{valid}}$, equivalently scoring each valid answer group by
$\max_{i\in I_a}e_i$~\citep{cobbe2021verifiers}.
Score-aware consensus pools evidence within a valid group using raw sums or
means, softmax-normalized weights, rank weights, or calibrated
log-odds~\citep{taubenfeld2025cisc,kang2025selfcertainty,kuang2026optimalaggregation}.
Every rule must declare deterministic tie handling and a fallback submitted
output when $\mathcal{I}_{\mathrm{valid}}=\emptyset$.

Other completed-bank reducers combine filtering, resampling, and gated selection. Offline DeepConf removes low-confidence traces; Majority-of-the-Bests bootstraps Best-of-$N$ decisions and returns their mode; and Best-of-Majority filters answer groups by frequency before reward-ranking the retained responses for Pass@$k$~\citep{fu2026deepconf,rakhsha2025majority,di2026bestofmajority}. When valid candidates exist, MBR selects
\[
i^\star\in\arg\max_{i\in\mathcal{I}_{\mathrm{valid}}}
\frac{1}{|\mathcal{I}_{\mathrm{valid}}|}
\sum_{j\in\mathcal{I}_{\mathrm{valid}}}[\mathbf{W}_n]_{ij},
\]
with the same all-invalid fallback~\citep{kumar2004mbr,freitag2022neuralmbr,bertsch2023mbr}.
Each method specifies a decision structure while leaving the evidence source separate. If the output contains both a canonical answer and a rationale, the protocol must also specify which trace accompanies the winning answer.

\paragraph{Score semantics are part of the algorithm.}
Naming the scorer does not fully specify a score-aware reducer. The protocol must state where scores come from, whether larger values are better, and how token- or step-level scores are reduced. Answer canonicalization and invalid-output handling determine which candidates enter the rule. Score transformations, temperatures, and filtering thresholds determine how evidence enters the decision. The protocol must also define ties and fallbacks, and identify any evaluator prompts or models. Calibration parameters and other tunable values must be fixed on held-out data, not test labels.

Order-only rules such as hard Best-of-$N$, rank-based filtering, rank voting, and Majority-of-the-Bests are invariant to strictly increasing score transformations that preserve ties. Rules based on raw sums or means, softmax weights, or log-odds are not. Sequence and mean log-probability impose different length preferences. Cross-question calibration need not predict a confidence score's usefulness for within-question aggregation~\citep{taubenfeld2025cisc}; calibrated log-odds may assign negative evidence to low-scoring candidates~\citep{kuang2026optimalaggregation}. With hard maximization, exposure to proxy-score errors increases as the bank grows~\citep{gao2023scaling,huang2025bestofn}.

\paragraph{Causal stopping.}
Generation cost can fall only when a causal controller stops future work; evidence extraction and fixed-bank selection act after their required traces are complete. Let
\[
\mathcal{H}_n
=
\sigma\!\left(x,(Y_i,a_i,m_i,e_i)_{i\le n},\mathbf{W}_n,\zeta_{0:n}\right)
\]
be the information available after $n$ completed rollouts, including all
revealed evidence, pairwise values, and controller random state $\zeta_{0:n}$.
An across-rollout rule is a bounded stopping time
$\tau_{\mathrm{roll}}\le N_{\max}$ satisfying
$\{\tau_{\mathrm{roll}}\le n\}\in\mathcal{H}_n$, and its final decision may use
only the bank observed through $\tau_{\mathrm{roll}}$. In plain terms, whether
the procedure has stopped by rollout $n$ may depend on the first $n$ completed
rollouts, but not on any later candidate or inference-time signal. Adaptive-Consistency
stops when the posterior probability that the current count leader has greater
latent mass than its runner-up crosses a threshold, whereas Early-Stopping
Self-Consistency stops after a valid unanimous answer window~\citep{aggarwal2023adaptive,li2024esc}.
The sample cap $N_{\max}$ is separate from the cumulative budget $B$: before
launching or continuing an operation, the controller must establish that its
declared worst-case charge keeps total cost at most $B$. We assume operations
that would overshoot are not launched; a protocol using causal truncation
instead must declare that terminal rule. Warm-up, discarded, evaluator,
controller, and final-decision work all consume the same declared budget unit.

For rollout $i$, let $\mathcal{F}_{i,t}$ contain the permitted prior history,
revealed evaluator/controller state, and the current rollout's information
through token $t$. A within-rollout stopping time $\tau_{\mathrm{tok}}$
satisfies $\{\tau_{\mathrm{tok}}\le t\}\in\mathcal{F}_{i,t}$: it cannot use the
hidden suffix or eventual answer to decide whether to truncate the current
trace. Truncation emits an explicit terminal outcome, which may parse as
invalid, rather than treating an unfinished prefix as a completed leaf.
DeepConf, for example, uses completed warm-up traces to calibrate a confidence
threshold and may terminate later traces when sliding-window token confidence
falls below it~\citep{fu2026deepconf}. Across-rollout rules can avoid only later
traces, whereas within-rollout rules can also avoid a suffix of the current
trace. Offline filtering of completed traces may change the returned answer but
saves no generation tokens.

\paragraph{Comparison and compute accounting.}
A \emph{shared-bank} comparison measures aggregation conditional on one completed bank. Each offline protocol receives the same candidates and only the inference-time signals it is permitted to use. Holding $E_\eta$, parsing, canonicalization, candidate eligibility, and score transformations fixed isolates $\mathcal{G}_\delta$; otherwise, those differences belong to the compared post-generation protocols. If both evidence and decision components vary, the comparison covers the full post-generation stage $(E_\eta,\mathcal{G}_\delta)$ under a fixed proposal and realized bank. Neither variant can estimate savings from adaptive stopping or changes caused by a different generation policy. Generalization beyond the realized bank still requires uncertainty estimates over prompts and candidate draws. An \emph{end-to-end} comparison instead allows each complete protocol to generate its own stream, acquire its own evidence, execute its stopping rule, and be evaluated by $M_B$.

For one declared additive work unit, the matched budget decomposes as
\[
C_{\mathrm{total}}
=
C_{\mathrm{gen}}+C_{\mathrm{eval}},
\qquad
C_{\mathrm{eval}}
=
C_{\mathrm{signal}}+C_{\mathrm{control}}+C_{\mathrm{decision}}.
\]
This accounting includes warm-up and discarded tokens, verifier or judge calls, repeated controller evaluations, and aggregation itself. Equal sample counts need not imply equal compute: external scoring adds candidate-wise cost, and naive pairwise MBR requires $O(n^2)$ utility evaluations~\citep{cheng2023cbp,jinnai2024ambr}. Latency under parallel or overlapping stages, throughput, peak memory, and other noncommensurate resources are not additive terms in this equation and should be reported separately alongside any scalar work budget. End-to-end utility should be plotted against total cost. Shared-bank results
instead identify differences attributable to the post-generation aggregation
stage, and holding evidence fixed narrows attribution to the decision rule. The discovery--stability profile in \Cref{ssec:tts_metrics} remains complementary because it describes candidate availability under the proposal, not the success probability of a particular aggregation protocol.

\subsection{Evaluation targets and metrics}
\label{ssec:tts_metrics}

Evaluation under test-time scaling should separate the performance of the
deployed inference system from the shape of the candidate bank it induces.

\paragraph{End-to-end system performance.}
Let \(\mathcal{P}\) denote the task distribution and
\(\mathcal{D}_{\mathrm{eval}}=\{x_q\}_{q=1}^Q\) the evaluation set. For task utility
\(U_x\), budgeted algorithm \(\mathcal{A}_B\), and algorithmic randomness \(\xi\),
\[
M_B
=
\mathbb{E}_{x\sim \mathcal{P},\xi}
\!\left[
U_x\!\left(\hat o_B(x;\xi)\right)
\right],
\qquad
\widehat M_B
=
\frac{1}{Q}\sum_{q=1}^Q
U_{x_q}\!\left(\hat o_B(x_q;\xi_q)\right).
\]
This is the estimand for capability claims: it is the scaling curve \(G(B)\)
of \Cref{sec:form:setup_taxonomy}, with the algorithmic randomness \(\xi\)
written explicitly. Pairwise win
rates, judge-aggregated rankings, and other relative comparisons define
separate estimands; when they are used, the judge prompt, response order,
reference material, and aggregation rule are part of the evaluation protocol.

\paragraph{Discovery--stability profile.}
For repeated-sampling diagnostics, fix the proposal or inference protocol that
produces completed candidates. Let \(Z_{q,i}\in\{0,1\}\) indicate whether
candidate \(i\) for prompt \(q\) is correct. Under the independent-attempt
abstraction, let \(p_q=\Pr(Z_{q,i}=1\mid x_q)\) be the latent single-attempt
success probability for prompt \(q\). For \(k\) fresh attempts,
\[
X_{q,k}\mid p_q \sim \mathrm{Binomial}(k,p_q).
\]
For threshold \(t\in\{1,\ldots,k\}\), define the binomial tail kernel
\[
\kappa_{k,t}(p)
=
\Pr\{\mathrm{Binomial}(k,p)\ge t\}
=
\sum_{j=t}^k \binom{k}{j}p^j(1-p)^{k-j}.
\]
The dataset-level discovery--stability profile at budget \(k\) is
\[
S_k=(S_{k,1},\ldots,S_{k,k}),
\qquad
S_{k,t}
=
\frac{1}{Q}\sum_{q=1}^Q \kappa_{k,t}(p_q).
\]
Thus \(S_{k,t}\) is the expected fraction of prompts on which a fresh
\(k\)-attempt evaluation produces at least \(t\) correct candidates. Low
thresholds measure discovery, high thresholds measure repeatability, and the
full profile records how quickly occasional success decays into stable success.

When an observed bank contains \(N\) candidates per prompt and \(k\le N\), let
\(\nu_q=\sum_{i=1}^N Z_{q,i}\). The corresponding finite-bank diagnostic is the
without-replacement tail
\[
\widehat S^{\mathrm{bank}}_{k,t}
=
\frac{1}{Q}\sum_{q=1}^Q
\sum_{j=t}^k
\frac{
\binom{\nu_q}{j}\binom{N-\nu_q}{k-j}
}{
\binom{N}{k}
},
\]
with invalid binomial coefficients interpreted as zero. This statistic
evaluates the threshold event for a size-\(k\) subset drawn without replacement
from the observed bank, whereas \(S_k\) is the prospective latent profile. For
banks produced by adaptive search or other
dependent procedures, the finite-bank profile remains a descriptive summary of
the returned bank; the latent i.i.d.\ interpretation should be invoked only when
it matches the sampling protocol.

A Bayesian report separates finite-bank evidence from the latent profile. With
\(p_q\sim\mathrm{Beta}(\alpha^0_q,\beta^0_q)\), defaulting to
\(\mathrm{Beta}(1,1)\) unless an auxiliary prior bank is declared, conjugacy
gives
\[
p_q\mid Z_{q,1:N}
\sim
\mathrm{Beta}(a_q,b_q),
\qquad
a_q=\alpha^0_q+\nu_q,\quad
b_q=\beta^0_q+N-\nu_q.
\]
The posterior mean of each profile coordinate is the beta-binomial predictive
tail
\[
\mu_{k,t}
=
\frac{1}{Q}\sum_{q=1}^Q
\sum_{j=t}^k
\binom{k}{j}
\frac{
\mathrm{B}(a_q+j,b_q+k-j)
}{
\mathrm{B}(a_q,b_q)
},
\qquad
t=1,\ldots,k ,
\]
where \(\mathrm{B}(\cdot,\cdot)\) is the beta function. Shared posterior draws
of \(p_{1:Q}\) should be used to evaluate all thresholds and all scalar
summaries, preserving their posterior dependence.

For rubric-valued outcomes, the same construction replaces the Bernoulli model
with a Dirichlet--categorical model, fixes a category score vector in advance,
and computes tail probabilities of the normalized \(k\)-sample rubric score.
The binary exact-match profile above is the two-category case with score vector
\((0,1)\).

\paragraph{Scalar views of the profile.}
The profile \(S_k\) is the primary bank-level object. A scalar metric is a
prespecified functional \(f(S_k)\), not a replacement for the profile. For a
linear threshold utility with weights
\(\omega_{t,k}\ge 0\) and \(\sum_{t=1}^k \omega_{t,k}=1\),
\[
U_\omega(k)
=
\sum_{t=1}^k \omega_{t,k}S_{k,t}.
\]
Common repeated-sampling metrics are coordinates or simple functionals of the
same profile, or are bounded by
them:
\[
\mathrm{Pass}@k = S_{k,1},
\qquad
\mathrm{pass}^{k}=S_{k,k},
\qquad
\mathrm{Maj}@k \ge S_{k,\lfloor k/2\rfloor+1}
\quad\text{(single canonical target)},
\]
\[
\mathrm{G\text{-}Pass}@k_\tau
=
S_{k,\lceil \tau k\rceil},
\qquad
0<\tau\le 1,
\]
\[
\mathrm{mG\text{-}Pass}@k
=
\frac{1}{k-\lfloor k/2\rfloor}
\sum_{t=\lfloor k/2\rfloor+1}^{k} S_{k,t},
\qquad
\mathrm{Geom}@k
=
\left(S_{k,1}S_{k,k}\right)^{1/2},
\]
where the averaged form of \(\mathrm{mG\text{-}Pass}@k\) coincides with the
definition of \citet{liu2025your} for even \(k\).
The finite-bank versions are obtained by replacing \(S_{k,t}\) with
\(\widehat S^{\mathrm{bank}}_{k,t}\); in particular,
\(\widehat{\mathrm{Pass}}@k=\widehat S^{\mathrm{bank}}_{k,1}\).
The strict-majority coordinate \(S_{k,\lfloor k/2\rfloor+1}\) lower-bounds
plurality voting: a strict majority of correct candidates guarantees that the
correct answer wins the vote, whereas plurality can also succeed with fewer
correct candidates when incorrect answers disagree.
Ordinary single-sample pass rate is \(S_{1,1}\), and for any \(k\) the
uniform-threshold area recovers the same quantity:
\[
\frac{1}{k}\sum_{t=1}^k S_{k,t}
=
\frac{1}{Q}\sum_{q=1}^Q p_q.
\]

These quantities characterize candidate availability under the proposal; they do
not give the success probability of an aggregation rule. Selection rules such as
self-consistency, verifier reranking, MBR, and judge-based selection must be
evaluated end-to-end through \(M_B\). The same applies to adaptive stopping,
which changes the realized bank. Under the complete protocol in
\Cref{ssec:aggregation}, \(S_k\) remains a complementary diagnostic of the
candidate bank and its discovery--stability shape.

\subsection{Reproducibility}
\label{subsec:eval_reproducibility}

Under test-time scaling, a candidate bank and its submitted output together form
one realization of a randomized inference protocol. \emph{Exact replay} reconstructs
the same candidates, inference-time signals, stopping and aggregation decisions,
and scores under fixed inputs, random streams, software, and hardware.
\emph{Distributional reproducibility} instead asks whether an independent rerun
returns a statistically compatible estimate of the same estimand, such as
$M_B$, the profile $S_k$, or a judge-defined win rate. A fixed seed can support
exact replay, but it does not measure variation over prompts, candidate draws,
judge calls, or numerical execution~\citep{bethard2022seeds}.

Candidate banks are not always independent. Shared prefix-search state,
verifier feedback, tool state, or controller decisions can couple candidates;
adaptive stopping can also make bank size depend on earlier outcomes. In these
cases, the latent i.i.d.\ interpretation of $S_k$ does not apply, although the
finite-bank profile remains descriptive (\Cref{ssec:tts_metrics}). A replay
record should include the raw candidates, inference-time signals, stopping and
aggregation decisions, and a random-stream identifier for each candidate.
Deriving each candidate's random state from the run, prompt, sample index, and
stream name prevents asynchronous batching from reassigning random variates
across candidates.

\paragraph{Quantifying variability.}
A distributional reproducibility claim should specify a compatibility criterion
and the random sources it covers. A prompt bootstrap over observed candidate
banks estimates variation due to prompt composition conditional on those banks;
it does not include new candidate draws, judge calls, or numerical
execution~\citep{efron1979bootstrap}. Claims that average over these sources
require independent protocol reruns or a resampling design that includes the
corresponding stages~\citep{bouthillier2021accounting,blackwell2024towards}. Each
replicate must recompute the reported bank statistic, reducer, or stopping rule
rather than resample an already aggregated score. Otherwise, uncertainty can be
narrower than rerun variability, and system rankings can change across
runs~\citep{miller2024adding,ye2024benchmarking,hariri2026dontpassk,hariri2026ranking}.

\paragraph{Numerical execution.}
Precision, kernels, and batching can change a near-tied next-token decision and
propagate the difference through a reasoning trace; even greedy decoding may
therefore fail to replay exactly~\citep{hochlehnert2025sober,tahmasivand2025lmfix,yuan2025give}. An
exact-replay report should specify the inference library and version, numerical
formats, quantization, kernels, and batch schedule. A matched FP32 or
higher-precision audit using the same prompts and random-stream assignments can
estimate sensitivity to numerical format, but does not establish deterministic
replay~\citep{yuan2025give}.

\section{Empirical Study and Trace Corpus}\label{sec:exp}

Our empirical study uses fixed response banks to measure how quickly additional
sampling reveals a correct candidate and how often practical reducers select
one. The public release retains the responses, outcomes, and recorded signals
for three collections, so new reducers can be evaluated without regenerating
those responses. The empirical study has four blocks spanning different tasks,
model rosters, sampling protocols, and levels of signal detail
(\Cref{tab:release-inventory}). We therefore report each block separately.

\begin{table}[!t]
\centering
\small
\setlength{\tabcolsep}{3.2pt}
\begin{tabularx}{\linewidth}{@{}lrrrrY@{}}
\toprule
Study block & Questions & Banks & \(N\) & Responses & Other recorded signals \\
\midrule
\mmlupro{} + \bbh
  & 18,543 & 27 & 1 & 500,661
  & Parsed answers \\
\makecell[l]{\aimeboth, \hmmt, and \brumo}
  & 120 & 20 & 80 & 192,000
  & Chosen-token log probability and rank; outcome-verifier score \\
\makecell[l]{\aimesix, \hmmtboth,\\\cmimc, and \smt}
  & 186 & 7 & 80 & 104,160
  & Top-20 token probabilities; reference-based and reference-free verifiers \\
\supergpqa{} (subset)
  & 3,600 & 4 & 80 & 1,152,000
  & Top-20 token probabilities and both verifier families \\
\midrule
\textbf{Total} & \textbf{22,449} & -- & -- & \textbf{1,948,821} & \\
\bottomrule
\end{tabularx}
\caption{\textbf{Response-bank inventory.} A bank is one model configuration
evaluated over a complete question set; \(N\) is the number of responses per
question in that bank. The signal-rich mathematics row includes three
independently generated medium-effort repeats used only for repeated-run
diagnostics.}
\label{tab:release-inventory}
\end{table}

\subsection{Broad evaluation across knowledge and symbolic reasoning}

This block covers 14 \mmlupro{} domains~\citep{wang2024mmlu} and the
23 \bigbenchhard{} tasks~\citep{suzgun2022challenging}. We evaluate 27
open-weight reasoning models with one zero-shot chain-of-thought response per
question, yielding 500,661 responses. Across the roster, \mmlupro{} exact match
ranges from 3.13\% to 70.47\%, and \bbh{} flexible-extraction exact match ranges
from 25.97\% to 82.20\%. We report the suites separately because their tasks,
answer spaces, and parsers differ.

\subsection{Candidate discovery outpaces answer selection}
\label{ssec:exp-repeated-math}

This block contains 80 seeded responses for every combination of 20 model
configurations and 120 questions from \aimefour, \aimefive, \hmmt, and
\brumo~\citep{aime2024dataset,aime2025dataset,
matharena_hmmt2025_dataset,matharena_brumo2025_dataset}, for 192,000 responses.
We compute exact candidate-bank statistics under without-replacement sampling
at \(k\in\{1,2,4,8,16,32,64,80\}\), and replay reducers on nested random
subsets of the same banks. Because every reducer sees subsets of the same
response banks, differences among reducers cannot be attributed to newly
generated candidates.

\begin{figure}[!t]
    \centering
    \includegraphics[width=0.95\linewidth]{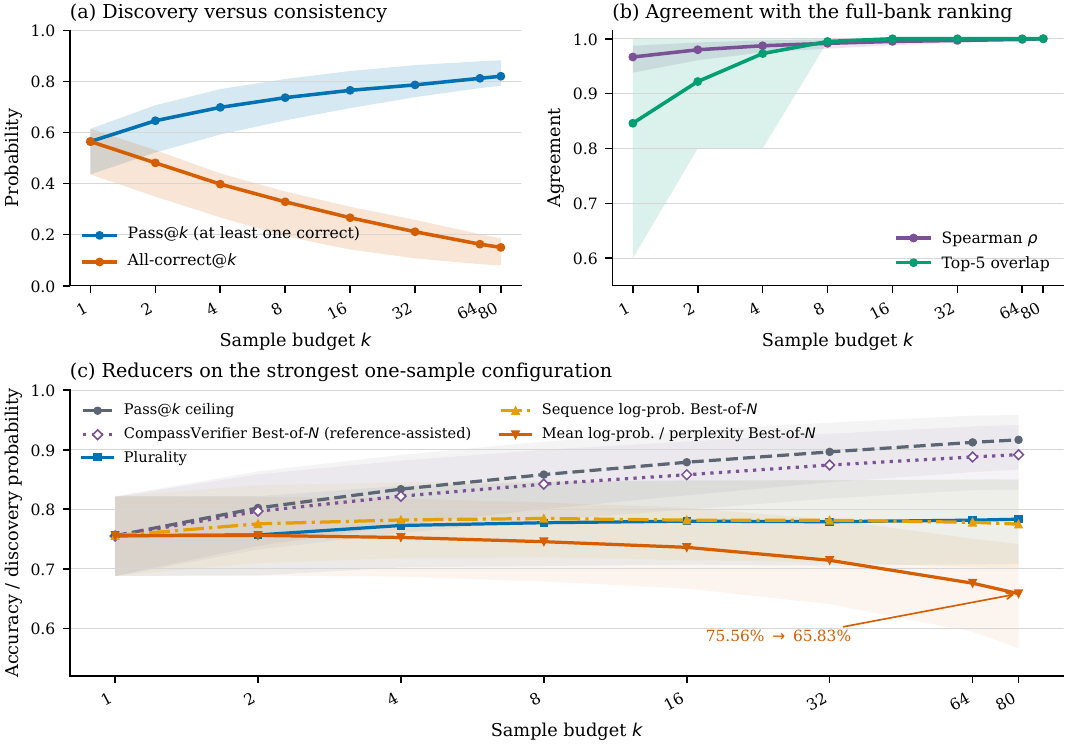}
    \caption{\textbf{Candidate availability and reducer accuracy under repeated
    sampling.} (\textbf{a}) Median exact Pass@\(k\) and
    \(\mathrm{pass}^{k}\) across the 20 configurations; bands span the
    interquartile range of configurations. (\textbf{b}) Mean agreement of
    subset rankings with the \(N=80\) ranking; bands are central 95\% ranges
    over 200 paired subset replays. (\textbf{c}) Reducers for
    Qwen3-30B-A3B-Thinking-2507, the strongest single-response configuration
    in this roster; bands are 95\% prompt-bootstrap intervals. CompassVerifier
    sees the reference answer and is shown only as a reference-assisted
    diagnostic. Mean log probability and negative perplexity induce the same
    ordering, so one curve represents both.}
    \label{fig:experiment2-scaling}
\end{figure}

Across the roster, the median Pass@\(k\) rises from 56.49\% at \(k=1\) to
82.08\% at \(k=80\), whereas the median all-correct coordinate falls to
15.00\%. For this roster, mean Spearman agreement with the \(N=80\) accuracy
ranking is 0.967 at \(k=1\) and 0.992 at \(k=8\). These values describe the 20
configurations and their sampled banks; they do not establish a general
small-sample guarantee.

Qwen3-30B-A3B-Thinking-2507 has the highest single-response accuracy in the
roster, at 75.56\%. At \(k=80\), Pass@80 is 91.67\% and the finite-bank
all-correct coordinate is 43.33\% (\Cref{fig:experiment2-scaling}, left).
Literal answer plurality reaches 78.33\%, and sequence-log-probability selection
reaches 77.50\%. Accuracy from selecting the response with the highest mean
token log probability instead falls from 75.56\% to 65.83\% as the bank grows
(\Cref{fig:experiment2-scaling}, right). For this score, expanding the candidate
bank from 1 to 80 therefore lowers submitted-answer accuracy despite raising
Pass@\(k\). The reference-assisted CompassVerifier diagnostic reaches 89.17\%,
2.50 percentage points below Pass@80, but it has access to the gold answer and
is not an inference-time result.

\subsection{Full trace collection on competition mathematics}
\label{ssec:exp-rich-math}

This block contains 186 problems from five competition datasets distributed by
MathArena~\citep{dekoninck2026matharena}: \aimesix,
\hmmtfebsix, \hmmtnovfive, \cmimc, and \smt~\citep{matharena_aime2026_dataset,
matharena_hmmtfeb2026_dataset,matharena_hmmtnov2025_dataset,
matharena_cmimc2025_dataset,matharena_smt2025_dataset}.

The effort comparison comprises one Qwen3.6-35B-A3B~\citep{qwen36_hf} bank and
three gpt-oss-20b banks at low, medium, and high reasoning effort. Every trace
records the full response, rule-based outcome, chosen token, and up to 20 token
alternatives at every generation position.

\begin{figure}[!t]
    \centering
    \includegraphics[width=\linewidth]{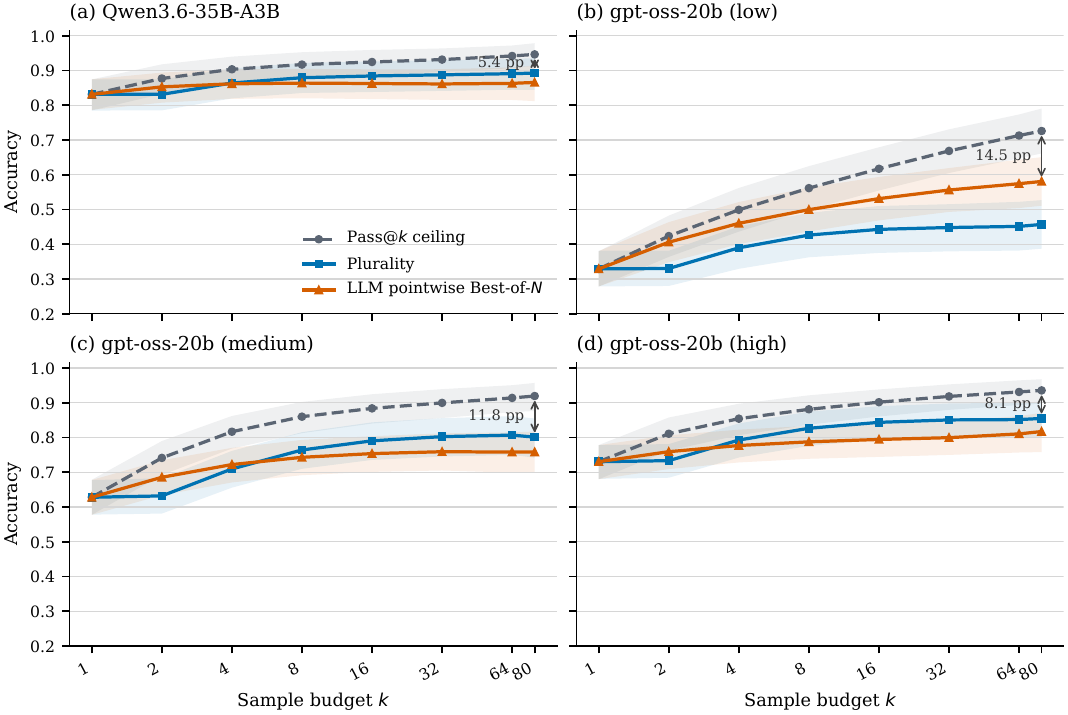}
    \caption{\textbf{Finite-bank scaling on the five 2025--2026 competition
    sets.} Each panel compares the exact Pass@\(k\) ceiling with
    literal answer plurality and pointwise Best-of-\(N\) using the
    reference-free verifier described below. Arrows mark the \(k=80\) gap
    between Pass@80 and the better of the two observed reducers. Shading gives
    95\% prompt-bootstrap intervals conditional on the observed 80-response
    banks. Intermediate reducer points average 2,000 nested subset replays;
    \(k=1\) and \(k=80\) use exact endpoints. The horizontal axis counts
    generated candidates and excludes verifier computation.}
    \label{fig:matharena-scaling}
\end{figure}

Pass@\(k\) increases for all four banks (\Cref{fig:matharena-scaling}). At
\(k=80\), Pass@80 is 94.62\% for Qwen3.6 and 72.58\%, 91.94\%, and 93.55\% for
gpt-oss low, medium, and high. The corresponding accuracies after reference-free
pointwise selection are 86.56\%, 58.06\%, 75.81\%, and 81.72\%. The remaining
8.06--16.13 percentage-point gaps measure the share of questions for which a
correct candidate is available but the reducer does not select it. Overlapping
prompt-bootstrap intervals and differences in response length preclude a total
ordering of the generators. Among high-effort responses, 2,399 of 14,880
(16.12\%) reach the common 81,920-token cap, compared with none at low effort.

\subsection{Reference-assisted and reference-free verifier signals}
\label{ssec:exp-verifiers}
We compute one reference-assisted score and one
reference-free score. CompassVerifier-3B (OpenCompass) receives the question,
reference answer, and candidate response and predicts A/B/C for correct,
incorrect, or invalid responses~\citep{CompassVerifier}. We retain both its
direct label distribution and a null-baseline-adjusted contextual distribution.
The reference-dependent score is an outcome diagnostic and cannot be used for
inference-time selection when gold answers are unavailable.

The reference-free score is a pointwise adaptation of the pairwise
LLM-as-a-Verifier method~\citep{kwok2026llmverifier}. Qwen3.6 receives one
problem and one candidate response, with no reference answer, ground truth,
Compass result, or other candidate. It scores problem understanding, reasoning
validity, and conclusion support on an ordinal A--T scale. We map A through T to
20 through 1, compute the expected score from the returned,
visible-mass-renormalized scoring-token probabilities, normalize each criterion
to \([0,1]\), and average the three values. The score can rank candidates without
a gold answer.

Against the rule-based outcome on 104,160 competition-mathematics responses,
the trace-level ROC AUC is 0.983 for direct Compass A, 0.977 for contextual
Compass A, and 0.871 for the pointwise score (\Cref{fig:verifier-diagnostics}).
The pointwise AUC is 0.744 on Qwen-generated responses and 0.876 on gpt-oss
responses.

\subsection{\supergpqa{}}

For \supergpqa~\citep{supergpqa2025}, we select 50 questions from each of 72
EvalScope fields, giving 3,600 questions~\citep{evalscope_2024}. The four banks
contain 80 responses per question from Qwen3.6 and gpt-oss at low, medium, and
high effort, for \(3{,}600\times4\times80=1{,}152{,}000\) responses.

For the gpt-oss high bank, mean response accuracy is 45.03\%, Pass@80 is
81.94\%, and \(\mathrm{pass}^{80}\) is 9.47\%. Of the 3,600 questions, 650 have
no correct response and 341 are answered correctly in all 80 responses.
Field-level response accuracy ranges from 21.33\% in Aquaculture to 76.78\% in
Mathematics (\Cref{fig:supergpqa-fields}).

\section{Conclusion}
\label{sec:conclusion}

Test-time scaling encompasses a family of budgeted inference algorithms. Viewing these algorithms as operations over the implicit prefix tree distinguishes among single-trajectory deliberation, leaf-level sampling with terminal reduction, and prefix-level search. It also identifies the proposal, evidence map, decision rule, and causal stopping controller as distinct components of the inference process. Accordingly, the complete inference system is the scientifically meaningful unit of comparison.

This perspective clarifies how reasoning systems should be evaluated. End-to-end utility and candidate-bank diagnostics answer different questions; shared-bank and end-to-end comparisons support different claims; compute accounting must include generation, evaluation, control, and decision; and uncertainty estimates must reflect the deployed inference protocol. Our framework also unifies repeated-sampling evaluation through the discovery--stability profile and distinguishes exact replay from distributional reproducibility. We analyze and publicly release 1,403,520 model attempts sampled from benchmarks covering broad knowledge, symbolic reasoning, and competition mathematics.

Our empirical study demonstrates the practical consequences of this system-level view. Model rankings change with the rubric, sampling budget, benchmark, and commitment standard; inference-time signals that predict correctness within individual trajectories need not reliably rank models; and strict answer extraction can conflate abstention with error. These results suggest that progress in test-time scaling should be assessed through reproducible inference systems with explicitly reported inference protocols, compute budgets, and uncertainties, rather than through checkpoint scores obtained under incompletely specified procedures.

\subsubsection*{Acknowledgments}
This research was supported in part by NSF awards 2117439, 2112606, and 2320952.

\bibliography{main}
\bibliographystyle{tmlr}
\clearpage

\appendix

\section{Notation}
\label{app:notations}

\noindent The formalization (\Cref{sec:form:tts}), evaluation framework (\Cref{sec:eval}), and leaf- and prefix-level variants in Appendices~\ref{app:leaf_reducer_variants} and~\ref{app:prefix_search_variants} use a common notation.

\vspace{0.5em}
\noindent\textbf{Core objects and budgeted inference.}

{\small
\renewcommand{\arraystretch}{1.08}
\begin{tabularx}{\linewidth}{@{}lY@{}}
\toprule
Symbol & Meaning \\
\midrule
$x \in \mathcal{X}$ & Input prompt or problem instance. \\
$p_\theta(\cdot\mid x)$ & Base autoregressive language model. \\
$\mathcal{T}(x)$ & Implicit rooted prefix tree induced by generation from $x$. \\
$\mathcal{L}(x)$ & Terminal leaves of $\mathcal{T}(x)$, i.e., completed generations. \\
$z \preceq y$ & Prefix relation: search state or prefix $z$ lies on the path to completed leaf $y$. \\
$\pi$ & Local generation policy or decoder used to extend a trajectory. \\
$q_\pi(y\mid x)$ & Proposal distribution over completed leaves induced by $p_\theta$ together with policy $\pi$. \\
$\mathcal{A}_B$ & Budgeted test-time algorithm with total cost at most $B$. \\
$\mathcal{B}$ & Declared set of allowed nonnegative budgets. \\
$o_t,\; c(o_t)$ & Primitive operation at step $t$ and its associated cost. \\
$B,\; b_t$ & Total test-time budget and remaining budget at step $t$. \\
$\mathcal{O}(x)$ & Disjoint union of completed-leaf and answer-valued output spaces. \\
$U_x(o)$ & Task utility of output $o\in\mathcal{O}(x)$ on instance $x$. \\
$\hat{o}_B(x)$ & Unified output returned by $\mathcal{A}_B$; written $\hat{y}_B(x)$ or $\hat{a}_B(x)$ according to its type. \\
$\xi$ & Internal (algorithmic) randomness of $\mathcal{A}_B$; $\xi_q$ is its realization on prompt $x_q$. \\
$G(B)$ & Test-time scaling curve $\mathbb{E}_{x\sim\mathcal{P},\,\xi}\bigl[U_x(\hat{o}_B(x;\xi))\bigr]$. \\
$\pi_{\mathrm{seq}},\; \nu_t$ & Single-trajectory controller and its meta-action at step $t$ (distinct from the correct-candidate count $\nu_q$). \\
\bottomrule
\end{tabularx}
}

\vspace{0.6em}
\Needspace{8\baselineskip}
\noindent\textbf{Output interpretation and evaluation signals.}

{\small
\renewcommand{\arraystretch}{1.08}
\begin{tabularx}{\linewidth}{@{}lY@{}}
\toprule
Symbol & Meaning \\
\midrule
$\mathrm{Parse}(y)=(r,a)$ & Deterministic parser that extracts reasoning trace $r$ and answer $a$ from completed generation $y$. \\
$\Psi(y)=(d,s)$ & Decomposition of $y$ into a deterministically verifiable component $d$ and a non-verifiable remainder $s$. \\
$V_P(x,d)$ & Programmatic verifier applied to instance $x$ and verifiable artifact $d$. \\
$\mathcal{V}$ & Codomain of verifier outputs, which may be Boolean, graded, diagnostic, or canonicalized. \\
$\mathrm{PASS}$ & Subset of verifier outputs treated as acceptable by a reducer. \\
$J_\phi(x,\omega)$ & Learned evaluator on object $\omega$, where $\omega$ may be a completed leaf $y$ or a partial state $z$. \\
\bottomrule
\end{tabularx}
}

\vspace{0.6em}
\Needspace{8\baselineskip}
\noindent\textbf{Decoding protocols and induced proposals.}

{\small
\renewcommand{\arraystretch}{1.08}
\begin{tabularx}{\linewidth}{@{}lY@{}}
\toprule
Symbol & Meaning \\
\midrule
$\Sigma$ & Token vocabulary. \\
$h_t,\; \ell_t,\; p_t$ & Partial generation $(x,y_{<t})$, its logits, and the softmax token probabilities. \\
$D_\lambda,\; \sigma_t,\; \chi_t$ & One-token decoder with configuration $\lambda$, its finite state, and side information. \\
$u_t,\; S_t$ & Nonnegative token measure and retained support of the energy--gate decoder (distinct from the MBR utility $u(\cdot,\cdot)$ and the prefix score $S(z)$). \\
$\widetilde{q}_t$ & One-token proposal $\operatorname{Norm}_{S_t}(u_t)$. \\
$q_\lambda(y\mid x)$ & Sequence-level proposal induced by composing the local kernels; written $q_\pi$ when the protocol is denoted $\pi$. \\
\bottomrule
\end{tabularx}
}

\vspace{0.6em}
\Needspace{8\baselineskip}
\noindent\textbf{Evaluation under test-time scaling.}

{\small
\renewcommand{\arraystretch}{1.08}
\begin{tabularx}{\linewidth}{@{}lY@{}}
\toprule
Symbol & Meaning \\
\midrule
$\mathcal{P}$ & Task distribution over problem instances. \\
$\mathcal{D}_{\mathrm{eval}}=\{x_q\}_{q=1}^{Q}$ & Evaluation set with $Q$ prompts or problem instances. \\
$M_B,\; \widehat{M}_B$ & Population and empirical end-to-end performance of algorithm $\mathcal{A}_B$ under budget $B$. \\
$Z_{q,i}$ & Binary correctness indicator for candidate $i$ on prompt $q$ inside a sampled bank. \\
$\nu_q$ & Number of correct candidates in the bank for prompt $q$, i.e., $\sum_{i=1}^{N} Z_{q,i}$. \\
$p_q$ & Latent single-attempt success probability for prompt $q$ under the fixed proposal. \\
$k,\; t$ & Attempt budget of a repeated-sampling diagnostic and success-count threshold, $1\le t\le k$; finite-bank versions require $k\le N$. \\
$\kappa_{k,t}(p)$ & Binomial tail kernel $\Pr\{\mathrm{Binomial}(k,p)\ge t\}$. \\
$S_{k,t}$ & Discovery--stability profile coordinate: expected fraction of prompts with at least $t$ correct candidates among $k$ fresh attempts. \\
$\widehat{S}^{\mathrm{bank}}_{k,t}$ & Finite-bank (without-replacement) estimate of the same threshold event computed from an $N$-candidate bank. \\
$\mathcal{H}_n,\; \mathcal{F}_{i,t}$ & Information available after $n$ completed rollouts and through token $t$ of rollout $i$, respectively. \\
$\tau_{\mathrm{roll}},\; \tau_{\mathrm{tok}}$ & Across-rollout and within-rollout stopping times. \\
\bottomrule
\end{tabularx}
}

\vspace{0.6em}
\Needspace{8\baselineskip}
\noindent\textbf{Leaf-level scaling.}

{\small
\renewcommand{\arraystretch}{1.08}
\begin{tabularx}{\linewidth}{@{}lY@{}}
\toprule
Symbol & Meaning \\
\midrule
$N$ & Number of completed candidates in a leaf bank. \\
$[N]$ & Candidate-index set $\{1,\ldots,N\}$. \\
$\mathcal{Y}_N(x)=\{Y_i\}_{i=1}^N$ & Multiset of completed candidates generated for input $x$. \\
$Y_i$ & $i$th completed candidate (leaf). \\
$a_i,\; m_i$ & Canonical answer parsed from $Y_i$ and its inference-time signals; $a_i=\bot$ denotes an invalid candidate when applicable. \\
$\mathcal{I}_{\mathrm{valid}}$ & Indices of reducer-eligible candidates, excluding $a_i=\bot$ for answer-valued tasks. \\
$v_i,\; j_i$ & Verifier output $V_P(x,d_i)$ and learned score $J_\phi(x,Y_i)$ for candidate $Y_i$. \\
$E_\eta,\; \mathcal{E},\; e_i$ & Evidence map, evidence space, and candidate evidence object $e_i=E_\eta(x,Y_i,m_i)$; $e_i$ is scalar for score-based rules. \\
$\mathbf{W}_n$ & Pairwise utility matrix over the first $n$ completed candidates. \\
$\mathcal{G}_\delta,\; \hat{o}_n$ & Fixed-bank decision rule and its returned leaf, answer, or output set. \\
$\mathcal{R}_N$ & Leaf-level reducer that maps a candidate bank to a final leaf or answer. \\
$T(x,Y_i,v_i,j_i)$ & Tie-breaking score used by verifier-constrained selection. \\
$i^\star$ & Index of the selected candidate when the reducer returns one sampled leaf. \\
$\hat{q}_N(a\mid x)$ & Empirical answer distribution induced by the leaf bank. \\
$\psi(\cdot)$ & Deterministic projection of a leaf used inside an MBR objective. \\
$u(\cdot,\cdot)$ & Utility or similarity function used by an MBR-style reducer. \\
$C_{\mathrm{gen}}(n),\; C_{\mathrm{eval}}(n)$ & Generation cost and inference-time scoring, control, and reduction cost for a bank of size $n$. \\
$C_{\mathrm{signal}},\; C_{\mathrm{control}},\; C_{\mathrm{decision}},\; C_{\mathrm{total}}$ & Evidence-acquisition, controller, decision, and total protocol cost. \\
\bottomrule
\end{tabularx}
}

\Needspace{24\baselineskip}
\noindent\textbf{Prefix-level scaling.}

{\small
\renewcommand{\arraystretch}{1.08}
\begin{tabularx}{\linewidth}{@{}lY@{}}
\toprule
Symbol & Meaning \\
\midrule
$z$ & Search state: a literal token prefix or a macro-prefix consisting of one or more reasoning steps. \\
$\mathrm{Succ}(z)$ & Allowed successor states obtained by expanding prefix $z$. \\
$\mathcal{F}_t$ & Active frontier of unfinished search states at step $t$. \\
$\mathcal{Y}_t$ & Bank of completed leaves accumulated by a prefix search up to step $t$. \\
$\rho$ & Rollout policy used to continue a prefix to completion. \\
$q_\rho(\cdot\mid x,z)$ & Distribution over completed leaves obtained by continuing from prefix $z$ with rollout policy $\rho$. \\
$Q^\star_\rho(z)$ & Continuation value of prefix $z$ under rollout policy $\rho$. \\
$\widehat{Q}(z)$ & Approximate continuation-value estimate used for pruning or prioritization. \\
$S(z)$ & Generic prefix score used by the search controller. \\
$S_{\mathrm{LL}}(z)$ & Likelihood-based prefix score. \\
$R_\phi,\; S_{\mathrm{PRM}}(z)$ & Process reward model (step-level scorer) and the prefix score it induces. \\
$H(x,z)$ & Partial check; sound for rejection when $H(x,z)=0$ implies $U_x(y)=0$ for all $y\succeq z$. \\
$\pi_{\mathrm{ctrl}},\; a_t$ & Search controller and its meta-action (expand, generate successors, allocate rollouts, prune, terminate). \\
$\widehat{Q}_{\mathrm{MC}}(z)$ & Monte Carlo estimate of continuation value using $M_{\mathrm{roll}}$ rollouts from prefix $z$. \\
$K$ & Beam width in beam search. \\
$M_{\mathrm{roll}}$ & Number of rollouts used when estimating $\widehat{Q}_{\mathrm{MC}}(z)$. \\
$n(z),\; n(z,a)$ & Visit counts for state $z$ and action $a$ in an MCTS-style search tree. \\
$q_{\mathrm{search},B}(y\mid x)$ & Effective leaf distribution induced by a prefix-level search algorithm under budget $B$. \\
$Q_{\mathrm{search},B}(\mathcal{Y}\mid x)$ & Joint law of an ordered bank or multiset returned by prefix search. \\
\bottomrule
\end{tabularx}
}

\section{Leaf-level reduction: additional derivations and variants}
\label{app:leaf_reducer_variants}

Under the leaf-level formalism of \Cref{sec:form:leaflevel}, self-consistency is an empirical MBR rule, and weighted candidate banks use the same template. Answer selection and rationale presentation remain separate decisions.

\subsection{Self-consistency as empirical MBR}

Let $\psi(y)$ be the extracted answer, i.e., $\psi(y)=a$ with
$(r,a)=\mathrm{Parse}(y)$, let
$\mathcal{I}_{\mathrm{valid}}=\{i\in[N]:a_i\ne\bot\}$, and suppose this set is
nonempty. With $u(a,a')=\mathbf{1}[a=a']$, the empirical MBR objective over the
eligible leaves of $\mathcal{Y}_N(x)=\{Y_i\}_{i=1}^N$ is
\[
\widehat{R}_N(i)
=\frac{1}{|\mathcal{I}_{\mathrm{valid}}|}
\sum_{k\in\mathcal{I}_{\mathrm{valid}}}u\!\left(a_i,a_k\right)
=\frac{1}{|\mathcal{I}_{\mathrm{valid}}|}
\sum_{k\in\mathcal{I}_{\mathrm{valid}}}\mathbf{1}[a_i=a_k]
=\hat{q}_N(a_i\mid x).
\]
Therefore,
\[
i^\star \in \arg\max_{i\in\mathcal{I}_{\mathrm{valid}}} \widehat{R}_N(i)
\qquad \Longleftrightarrow \qquad
\hat{a} \in \arg\max_{a\ne\bot} \hat{q}_N(a\mid x).
\]
Plurality self-consistency is thus exactly empirical MBR with a zero-one
answer-agreement utility~\citep{wang2022selfconsistency,kumar2004mbr}. The
declared all-invalid fallback and deterministic tie rule from
\Cref{sec:form:leaflevel} complete both definitions.

\subsection{Weighted banks and heterogeneous proposals}

A weighted variant replaces the uniform empirical distribution by
\[
\hat{q}_w(a\mid x)
=\frac{\sum_{i\in\mathcal{I}_{\mathrm{valid}}} w_i\,\mathbf{1}[a_i=a]}
{\sum_{i\in\mathcal{I}_{\mathrm{valid}}} w_i},
\qquad w_i\ge 0,\quad
\sum_{i\in\mathcal{I}_{\mathrm{valid}}}w_i>0.
\]
The corresponding weighted MBR objective is
\[
\widehat{R}_w(i)=\sum_{k\in\mathcal{I}_{\mathrm{valid}}}
\bar{w}_k\,u\!\left(\psi(Y_i),\psi(Y_k)\right),
\qquad
\bar{w}_k=\frac{w_k}{\sum_{j\in\mathcal{I}_{\mathrm{valid}}}w_j}.
\]
The weighting scheme applies to confidence-weighted consensus and to banks built from heterogeneous proposal distributions. If candidate $Y_i$ is drawn from proposal $q_i(\cdot\mid x)$ but the desired risk is defined under target proposal $q_\star(\cdot\mid x)$, assume $q_\star(\cdot\mid x)$ is absolutely continuous with respect to every contributing $q_i(\cdot\mid x)$. Then importance weights
\[
w_i \propto \frac{q_\star(Y_i\mid x)}{q_i(Y_i\mid x)}
\]
define a self-normalized importance estimator when the probability ratios are
available. If truncated decoding or adaptive search prevents their evaluation,
replacement weights are heuristic rather than an exact correction.

\subsection{Beyond exact-match consensus}

Open-ended outputs may not admit a canonical exact-match representation. A
reducer can instead use a utility or similarity kernel over projected
candidates,
\[
s_i = \sum_{\ell=1}^N u\!\left(\psi(Y_i),\psi(Y_\ell)\right),
\qquad
i^\star \in \arg\max_i s_i.
\]
Neural MBR instantiates this template with learned reference-based metrics \citep{freitag2022neuralmbr}, while Universal Self-Consistency uses an LLM-mediated notion of consistency to extend self-consistency beyond exact answer extraction \citep{chen2023usc}.

\subsection{Proxy misspecification and regularized selection}

When a reducer scores candidates using an imperfect proxy $j_i$, hard argmax selection can overoptimize proxy error as the bank grows \citep{gao2023scaling}. One stochastic alternative replaces hard argmax selection with
\[
\Pr(\hat{y}=Y_i\mid x,\mathcal{Y}_N)
\propto \exp\!\left(\alpha j_i\right),
\qquad \alpha\ge 0,
\]
which is uniform at $\alpha=0$ and converges as $\alpha\to\infty$ to uniform
selection over the tied maximum-score candidates. More
generally, one can combine a proxy score with Bayes-risk
regularization~\citep{jinnai2024rbon}, or soften the hard argmax into stochastic
selection~\citep{verdun2025softbon}, so that selection is not determined by the
single largest proxy score alone. Because the reducer is part of the inference
system being evaluated, changing the strength or regularization of selection
changes the object under comparison.

\subsection{Answer selection versus rationale presentation}

When benchmark utility depends only on the final answer, answer selection and rationale presentation are distinct decisions:
\[
\hat{a}\in\arg\max_{a\ne\bot} \hat{q}_N(a\mid x),
\qquad
i^\star \in \arg\max_{i:\, a_i=\hat{a}} \tilde{T}(x,Y_i).
\]
The first rule selects the returned answer. Conditional on that answer, the
second uses a rationale-quality score $\tilde{T}$ (for example, the learned
score $j_i$) to choose a sampled rationale for presentation. Both argmax
operations use the protocol's fixed tie rule; the all-invalid case uses its
declared fallback.
Keeping these decisions separate
prevents silent reversion to full-string MAP selection. In neural machine
translation, for example, MAP decoding can be a poor output-quality
rule~\citep{eikema2020map}.

\section{Prefix-level search: controller and evaluator variants}
\label{app:prefix_search_variants}

Prefix search allocates computation using scores on unfinished states. Its
evaluator and controller are separate design choices; together, they determine
the proposal distribution over leaves. A completed-leaf count also omits
computation spent on prefix scoring and tree maintenance.

\subsection{Evaluator--controller factorization}

A prefix-level search procedure can be written as a pair $(S,\pi_{\mathrm{ctrl}})$, where $S$ maps a partial state to a score or summary statistic and $\pi_{\mathrm{ctrl}}$ uses those summaries to choose meta-actions,
\[
a_t \sim \pi_{\mathrm{ctrl}}(\cdot\mid x,\mathcal{F}_t,\mathcal{Y}_t,b_t).
\]
The action $a_t$ may specify which node to expand, how many successors to
generate, how many rollouts to allocate, which states to prune, or when to
terminate. Classical MCTS-style procedures~\citep{kocsis2006uct,browne2012mcts},
Tree-of-Thoughts-style search~\citep{yao2023tot}, and RAP~\citep{hao2023rap}
all instantiate this evaluator--controller pattern with different scorers and
control rules; beam and best-first search fit the same factorization. A
factorial ablation over $S$ and $\pi_{\mathrm{ctrl}}$ can separate gains due to
local evaluation, compute allocation, and their interaction.

\subsection{Search-induced proposal shift}

Any controller together with its scorer and stopping rule induces an effective leaf distribution
\[
q_{\mathrm{search},B}(y\mid x)=\Pr\bigl(\hat{y}_B(x)=y\bigr),
\]
which generally differs from the decoder-induced proposal $q_\pi(y\mid x)$.
When search returns an ordered bank or multiset $\mathcal{Y}$, the corresponding
object is its joint law $Q_{\mathrm{search},B}(\mathcal{Y}\mid x)$
(\Cref{sec:form:prefixlevel}); per-leaf inclusion probabilities are only
marginals of this law. Prefix search therefore changes both compute allocation
and the proposal over leaves. The dependence structure and leaf frequencies of
a returned bank reflect the controller's branch-selection decisions. Threshold
metrics such as $\mathrm{Pass}@k$ and the profile
coordinates $\widehat{S}^{\mathrm{bank}}_{k,t}$ (\Cref{ssec:tts_metrics}) can
still be computed on such banks, but they should be interpreted as descriptive
summaries of the search-induced bank unless the experimental protocol
explicitly averages over repeated executions of the full search algorithm.

When $S$ is a learned reward or value model, prefix search is susceptible to
proxy misspecification in reward-model selection~\citep{gao2023scaling}.
Reward-guided tree search may prune or replace low-scoring
states~\citep{hung2025rewardguidedtsearch}. A branch that would yield a
high-utility leaf can therefore be eliminated before that leaf is generated.
This failure cannot occur in pure leaf-level reranking after the bank has been
generated.

\subsection{Compute accounting}

Prefix search incurs compute costs for partial-state scoring, rollouts, and tree
maintenance, in addition to final generations. Reporting only the number of
completed leaves can therefore be misleading: two methods may return the same
number of leaves while using very different amounts of model computation and
evaluator computation. At minimum, empirical comparisons should separate
generated-token cost from evaluator cost in a declared additive work unit, and
report latency, throughput, peak memory, and noncommensurate calls separately.
Comparisons should also report whether the implementation supports
shared-prefix execution. Tree-structured kernels can reuse key--value caches
across prefixes, changing the wall-clock frontier for search-based
methods~\citep{yao2025deft}.

\section{Reasoning LLMs and Evaluated Models}
\label{app:reasoning_llms}

Open-weight \emph{reasoning models} are trained or configured to emit
intermediate steps for multi-step tasks in mathematics, science, and
code~\citep{guo2025deepseek,qwen3technicalreport,abdin2025phi4reasoning}.
Their post-training methods use explicit reasoning traces, verifiable rewards,
or preference signals, while inference interfaces determine how reasoning is
elicited and budgeted. These choices affect reasoning accuracy, output length,
and inference cost. \Cref{fig:reasoning-systems-chronicle} presents a selective
chronology of the model families.

\subsection{Common post-training pipeline archetypes}
\label{app:reasoning_landscape}

Open-weight reasoning pipelines recur in three post-training archetypes.

\noindent\textbf{(1) SFT$\rightarrow$RL pipelines.}
A common recipe first applies supervised fine-tuning (SFT) on curated
chain-of-thought (CoT) demonstrations~\citep{wei2022cot}, then uses
reinforcement learning (RL) to optimize behavior under a reward signal. The
reward may reflect verifiable correctness, human or model preferences, or
conciseness. Examples include \textsc{Sky-T1-7B}~\citep{sky-t1-7b},
\textsc{Light-R1}~\citep{wen2025lightr1}, and Phi-4 reasoning
variants~\citep{abdin2025phi4reasoning}.

Adjacent designs use SFT-only distillation in the original
\textsc{Sky-T1}~\citep{sky-t1-32b-pre}, preference optimization in
\textsc{Sky-T1-Flash}~\citep{sky-32-flash}, or branch--merge distillation in
\textsc{TinyR1}~\citep{sun2025tinyr1}.

\noindent\textbf{(2) Distillation-based pipelines.}
Distillation uses teacher-generated reasoning traces as SFT data for a smaller
student. DeepSeek-R1 supplies traces for its distilled
students~\citep{guo2025deepseek}, while OpenThoughts3 uses QwQ-32B as its trace
generator~\citep{openthoughts3_12m}. Once the traces are fixed, the student is
trained with cross-entropy SFT rather than an online RL stage. Workflows such as
\textsc{Bespoke-Stratos}~\citep{bespoke_stratos},
\textsc{OpenThoughts/OpenThinker}~\citep{openthoughts}, and
\textsc{Open-R1}~\citep{openr1_repo, openr1_distill_7b} differ in both data
volume (from $\sim$10K to $>1$M examples) and filtering procedures such as
exact-match checks and unit tests.

\noindent\textbf{(3) Parameter-space fusion and merging.}
A third line omits gradient updates during the merge stage and combines aligned
pretrained or post-trained checkpoints directly in parameter space.
\emph{Select--Calculate--Erase} (SCE), for example, computes a structured
weighted merge of aligned checkpoint parameters~\citep{wan2025fusechat}.
Related FuseAI work includes distribution-level knowledge
fusion~\citep{wan2024knowledge}, the fuse-then-merge pipeline of
\citet{wan2025fusechat}, and SFT+DPO implicit fusion~\citep{yang2025fusechat};
the \textsc{FuseO1} family also uses SCE. Branch--merge schemes appear in
\textsc{TinyR1}~\citep{sun2025tinyr1}. The merge stage requires no new training
data, but it is restricted to compatible checkpoints.

\subsection{Cross-cutting mechanisms and inference controls}
\label{app:reasoning_controls}

Four mechanism classes distinguish the systems in
\Cref{fig:reasoning-systems-chronicle} and
Table~\ref{tab:reasoning-models-1}: verified SFT and trace distillation, RL and
preference optimization, inference-time control and aggregation, and
parameter-space composition.

\noindent\textbf{CoT distillation and verified SFT.}
CoT distillation transfers teacher-generated solution decompositions to a
student through standard SFT. Reasoning pipelines may filter these traces with
exact-match checks or unit tests before training~\citep{sky-t1-32b-pre,
bespoke_stratos,openthoughts}. When supervision includes human labels,
annotator reliability is a separate source of error~\citep{ghiasvand2026realm}.

\noindent\textbf{RL and preference optimization.}
RL-style post-training can optimize policies directly for correctness or
preference-aligned behavior. In open-weight work, it appears as outcome-based
RL in verifiable domains such as mathematics and code, or as preference
optimization that penalizes unnecessarily long reasoning. Direct
Preference Optimization (DPO) optimizes a preference objective without fitting
the explicit reward model and value function used in Reinforcement Learning
from Human Feedback (RLHF)~\citep{rafailov2023direct,ouyang2022instructgpt}.
Reasoning-model objectives may prefer shorter correct chains or penalize
redundant verification, as in \textsc{Sky-T1-Flash}~\citep{sky-32-flash}.

\noindent\textbf{Inference-time scaling and aggregation.}
Test-time scaling methods allocate inference-time compute by sampling and
aggregating multiple candidates, as in majority-vote
self-consistency~\citep{wang2022selfconsistency}. Other controls set a
``thinking budget'' that extends or truncates deliberation, trading token use
and latency against accuracy~\citep{muennighoff2025s1}. These procedures modify
the decision rule beyond the base conditional distribution and are therefore
part of the evaluated system.

\noindent\textbf{Model merging and fusion.}
Parameter-space fusion combines checkpoints without gradient updates. FuseO1
checkpoints~\citep{fuseo1_32b_preview,fuseo1_flash_32b_preview} use SCE merging
to combine long-CoT and concise-answering models, and the merge itself requires
no additional training data. Fusion can change reasoning length and response
style while holding the backbone family and parameter count fixed.

\subsection{Comparison of representative open-weight reasoning models}
\label{app:reasoning_model_comparison}

Table~\ref{tab:reasoning-models-1} groups evaluated families and related
reference models by post-training mechanism, training-data scale, base model,
and inference controls. Because sampling, aggregation, and reasoning budgets
can change performance without changing weights, benchmark reports must
identify both the checkpoint and the inference protocol.

The mechanisms also impose different dependencies. Verified SFT depends on
trace quality and filtering, while RL depends on reward construction and
verification. Inference-time controls determine generation and evaluation cost.
Parameter-space fusion instead requires compatible checkpoints and can change
response length and style without changing the parameter count.

\begingroup
\footnotesize
\renewcommand{\arraystretch}{1.25}
\setlength{\LTleft}{0pt}
\setlength{\LTright}{0pt}
\begin{longtable}{@{}
  >{\raggedright\arraybackslash}p{2.2cm}
  >{\raggedright\arraybackslash}p{2.0cm}
  >{\raggedright\arraybackslash}p{2.6cm}
  >{\raggedright\arraybackslash}p{2.4cm}
  >{\raggedright\arraybackslash}p{4.8cm}
  @{}
}
\caption{Family-level comparison of representative open-weight reasoning models.}
\label{tab:reasoning-models-1}\\
\toprule
\textbf{Model / Family} &
\textbf{Paradigm} &
\textbf{Training data} &
\textbf{Base model} &
\textbf{Technical remarks} \\
\midrule
\endfirsthead

\caption[]{Family-level comparison of representative open-weight reasoning models (continued).}\\
\toprule
\textbf{Model / Family} &
\textbf{Paradigm} &
\textbf{Training data} &
\textbf{Base model} &
\textbf{Technical remarks} \\
\midrule
\endhead

\bottomrule
\endlastfoot

DeepSeek-R1 (teacher) &
SFT + RL (verifiable) &
Multi-stage (cold-start + RL) &
DeepSeek-V3-Base &
R1-Zero applies RL from a base model; R1 adds cold-start data and multi-stage SFT/RL; teacher for open distillations~\citep{guo2025deepseek}. \\

DeepSeek-R1-Distill\(^{\dagger}\) &
Distillation (supervised fine-tuning) &
$\sim$800k supervised samples ($\sim$600k reasoning) &
Qwen2.5 (1.5B--32B), Llama-3.1-8B &
Teacher-generated supervised fine-tuning into standard backbones across small and midsize models~\citep{guo2025deepseek,deepseek_r1_distill_qwen32b_hf}. \\

QwQ-32B &
RL reasoning &
RL post-training (blog-reported) &
Qwen2.5-32B &
Dense 32B RL reasoner; reference ``thinking'' model used as teacher in multiple open pipelines~\citep{qwq32b,QwQHF}. \\

Qwen3 Thinking\(^{\dagger}\) (4B / 30B-A3B / Next-80B-A3B) &
SFT + RL (thinking) &
Long-CoT cold start + reasoning RL + mode fusion + distillation~\citep{qwen3technicalreport} &
Qwen3; sparse MoE variants &
Thinking-specialized checkpoints; 4B and 30B variants report 262K context; Next-80B uses high-sparsity MoE + hybrid attention with GSPO for RL stability~\citep{qwen3_4b_thinking_hf,qwen3_30b_a3b_thinking_hf,qwen3_next_80b_hf,zheng2025gspo}. \\

s1 / s1.1 (SimpleScaling) &
SFT-only (data-minimized) + inference-time control &
1k curated, math-dominant CoTs &
Qwen2.5-Instruct (multiple sizes) &
Minimal SFT with termination control; compatible with sampling and aggregation protocols~\citep{muennighoff2025s1}. \\

Sky-T1\(^{\dagger}\) (NovaSky) &
SFT + preference/RL &
$\sim$17k curated prompts + preference pairs (Flash) &
Qwen2.5-7B/32B &
Rejection-sampled and verified traces; Flash uses length-sensitive preference optimization (SimPO-style) + optional rewriting (FCS+1) to reduce overthinking~\citep{li2025skyt1,sky-t1-7b,sky-32-flash}. \\

Bespoke-Stratos\(^{\dagger}\) &
Distillation (verified SFT) &
17k verified CoTs (math/code/sci) &
Students: Qwen2.5-7B/32B / Teacher: DeepSeek-R1 &
Correctness-filtered distillation (exact-match math, unit tests for code); open Curator pipeline~\citep{bespoke_stratos}. \\

OpenThinker\(^{\dagger}\) (OpenThoughts) &
SFT-only (curated) &
114k--1.2M reasoning traces (math/code/sci/puzzles) &
Qwen2.5 (1.5B--32B) &
Early releases test answer verification; OpenThoughts3 reports 1,000+ recipe experiments; Evalchemy evaluation~\citep{guha2025openthoughts}. \\

Open-R1\(^{\dagger}\) (Distill) &
SFT-only (verified) &
350k verified traces (math/code/sci) &
Qwen2.5-Math-7B &
Mixture-of-Thoughts: verified DeepSeek-R1 traces; fixed checkpoint for scaling and verifiable-RL follow-ons~\citep{openr1_repo,openr1_distill_7b}. \\

Open-R1 (Math) / OlympicCoder &
SFT-only (verified/partial) &
220k math (verified); CodeForces-CoTs (\(\sim\)100k traces; partial) &
Qwen2.5-Math-7B; Qwen2.5-Coder-7B/32B &
Math Verify with a judge fallback for 12\% of math samples; code traces use public tests and are not exhaustively verified; decontamination is documented~\citep{openr1_math_220k,codeforces_cots,openr1_update3}. \\

LIMO / LIMO-v2 / LIMR\(^{\dagger}\) (GAIR) &
SFT-only (data-minimized) + RL (verifiable) &
800 curated math; 1.4k RL-selected items &
Qwen2.5-32B-Instruct (LIMO); Qwen2.5-Math-7B (LIMR) &
Small-data SFT and learning-impact selection for RL; v2 pairs an updated 800-example dataset with a 32B checkpoint~\citep{ye2025limo,gair_limo_v2_modelcard,gair_limo_v2_dataset,li2025limr}. \\

Light-R1 / TinyR1\(^{\dagger}\) (Qihoo360) &
SFT + preference/RL; merge (branch--merge) &
76k + 3k CoTs; verified preference pairs &
Qwen2.5; DeepSeek-R1-Distill backbones &
Curriculum/step-wise SFT + DPO; GRPO on distilled variants; TinyR1 uses branch specialization followed by merging across domains~\citep{wen2025lightr1,sun2025tinyr1}. \\

FuseO1\(^{\dagger}\) (FuseAI) &
Merge (parameter fusion) &
N/A &
DeepSeek-R1-Distill-32B, QwQ-32B-Preview, Sky-T1-32B (incl.\ Flash) &
SCE-style parameter-space merging of long-CoT and concise models; requires no new data during merging~\citep{fuseo1_repo}. \\

gpt-oss-20b\(^{\dagger}\) (OpenAI) &
CoT RL (MoE) &
CoT RL post-training (model card) &
OpenAI MoE (21B total / 3.6B active) &
MXFP4 quantization; low, medium, and high reasoning-effort settings vary token use, latency, and accuracy~\citep{openai2025gptoss120bgptoss20bmodel}. \\

Phi-4 Reasoning\(^{\dagger}\) (Microsoft) &
SFT + RL (verifiable) &
Distilled traces + \(\sim\)6k RL math &
Phi-4 (14B) &
Teachable-prompt selection + long-CoT SFT; outcome-based RL in ``plus''; evaluated with inference-time budget and sampling controls~\citep{abdin2025phi4reasoning}. \\

Nemotron Reasoning\(^{\dagger}\) (NVIDIA) &
SFT-only (OpenReasoning); SFT + verifiable RL (AceReason) &
Large-scale synthetic post-training &
Qwen2.5 (OpenReasoning, AceReason); hybrid Mamba--Transformer (Nano v2) &
OpenReasoning is SFT-only; AceReason uses SFT followed by verifiable RL; Nano v2 uses a hybrid Mamba--Transformer architecture~\citep{openreasoning_nemotron_1_5b_hf,liu2025acereason,nvidia2025nemotronnano2}. \\

EXAONE-4.0\(^{\dagger}\) (LG AI) &
SFT + preference/RL &
Multi-domain math/code/science mixtures &
EXAONE-4.0-32B / 1.2B &
Unified modes (chat vs.\ reasoning); long-context support; preference/RL stages to balance correctness and verbosity~\citep{exaone-4.0}. \\
\end{longtable}
\noindent\(\dagger\) At least one checkpoint or configuration from this family appears in the 20-configuration repeated-mathematics roster in Section~\ref{app:exp-math20}. Unmarked rows provide broader field context for that roster.\par
\endgroup

\subsubsection{DeepSeek-R1 and distilled students: RL post-training and supervised distillation}

DeepSeek-R1 uses large-scale RL on tasks with verifiable outcomes.
\textbf{DeepSeek-R1-Zero} is trained with RL directly from a base model, without
an SFT warm start. The report describes self-verification and reflection in its
outputs, along with lower readability and language mixing. \textbf{DeepSeek-R1}
adds cold-start data, SFT, RL, and rejection sampling to address those output
problems while retaining the reasoning behavior~\citep{guo2025deepseek}.

The \textbf{DeepSeek-R1-Distill} family transfers R1 outputs into standard
dense backbones rather than running RL directly on smaller policies. The
released students include Qwen2.5-based checkpoints at 1.5B, 7B, 14B, and 32B,
and a Llama-3.1-based 8B checkpoint~\citep{guo2025deepseek}. The Hugging Face
model cards report that the Qwen2.5 students are fine-tuned on \(\sim\)800K
DeepSeek-R1-curated samples~\citep{deepseek_r1_distill_qwen32b_hf}. The project
reports that these distilled checkpoints outperform its attempts to train
smaller models directly with RL~\citep{deepseek_r1_github}.

\subsubsection{Qwen reasoning models: thinking-mode post-training and sparse MoE}

Qwen treats ``thinking'' as an explicit post-training target and uses both dense
and sparse-MoE architectures. \textbf{QwQ-32B} is a dense 32B model post-trained
with RL; Qwen reports benchmark results comparable to DeepSeek-R1 and
o1-mini~\citep{qwq32b,QwQHF}. Unlike the distilled families above, QwQ-32B is
post-trained directly for reasoning.

\textbf{Qwen3} separates \emph{thinking} and \emph{non-thinking} behavior through
a staged pipeline: long-CoT cold start, reasoning RL, ``thinking mode fusion,''
general RL, and strong-to-weak distillation~\citep{qwen3technicalreport}. The
evaluated \textbf{Qwen3-4B-Thinking-2507} checkpoint has no mode toggle and a
reported 262K context window~\citep{qwen3_4b_thinking_hf}.
\textbf{Qwen3-30B-A3B-Thinking-2507} uses sparse MoE, with 30.5B total and 3.3B
active parameters (128 experts, 8 active), and reports the same context
length~\citep{qwen3_30b_a3b_thinking_hf}.

Outside the evaluated roster, \textbf{Qwen3-Next-80B-A3B-Thinking} combines
hybrid attention with a higher-sparsity MoE. Its model card attributes RL
stability and efficiency in this setting to Group Sequence Policy Optimization
(GSPO)~\citep{qwen3_next_80b_hf,zheng2025gspo}. The checkpoint's default prompt
template enforces thinking mode.

\subsubsection{NovaSky: verified trace distillation and length control}

The Sky-T1 line uses small verified corpora to train long-CoT behavior in
instruction-tuned backbones; later variants add preference optimization or
RL~\citep{sky-t1-32b-pre,sky-32-flash,sky-t1-7b}.

\textbf{Sky-T1-32B-Preview} fine-tunes
\textsc{Qwen2.5-32B-Instruct} on roughly 17K demonstrations distilled from
\textsc{QwQ-32B-Preview} and selected by rejection sampling. The reported math
and code results improve over the base model.
In the paper's perturbation experiments, shuffling or deleting reasoning steps
causes larger accuracy losses than perturbing surface
tokens~\citep{li2025skyt1,sky-t1-32b-pre}.

\textbf{Sky-T1-32B-Flash}~\citep{sky-32-flash} penalizes overthinking during
training. It forms preference pairs from the shortest and longest correct
candidates and adds hard-negative pairs that contrast a short incorrect answer
with a long correct one on difficult prompts. The method then applies a
length-sensitive, SimPO-style preference objective~\citep{meng2024simpo}. An
optional ``FCS+1'' rewrite retains the first correct mathematics solution and
one additional solution; code outputs are not rewritten.

\textbf{Sky-T1-7B}~\citep{sky-t1-7b} uses an iterative SFT/RL schedule starting
from \textsc{Qwen2.5-Math-7B}. It first applies SFT on a small verified set of
QwQ-distilled traces, followed by RL with PRIME-style prompt filtering and
rollouts on Eurus-2-RL-Data~\citep{cui2025prime}. A second distillation and SFT
stage precedes the final RLOO-based RL stage~\citep{ahmadian2024backtobasics}.

\subsubsection{Bespoke-Stratos: reasoning distillation as data curation}

Bespoke-Stratos uses verified teacher traces as cross-entropy SFT data. The
pipeline distills \textsc{DeepSeek-R1} outputs into
\textsc{Bespoke-Stratos-17k} and fine-tunes \textsc{Qwen2.5-Instruct}
backbones at multiple scales~\citep{bespoke_stratos}.

\textbf{Bespoke-Stratos-17k} contains 16.7K examples produced with a
Sky-T1-style distillation recipe~\citep{bespoke_stratos17k}. Candidate traces
from \textsc{DeepSeek-R1} are filtered by strict answer checks for mathematics
and unit-test execution for code before SFT.

The project releases two checkpoints. \textbf{Bespoke-Stratos-32B} fine-tunes
\textsc{Qwen2.5-32B-Instruct} on this dataset, and
\textbf{Bespoke-Stratos-7B} applies the same procedure to
\textsc{Qwen2.5-7B-Instruct}~\citep{bespoke_stratos32b,bespoke_stratos7b}. The
project reports improvements over both base models on its mathematics and code
evaluations~\citep{bespoke_stratos}.

\subsubsection{OpenThoughts/OpenThinker: scaling supervised reasoning data}

OpenThoughts studies SFT-only reasoning post-training with composed synthetic
traces, testing how performance changes with the volume and composition of the
supervision data~\citep{guha2025openthoughts,openthoughts}.

The first release, \textbf{OpenThoughts-114K}~\citep{openthoughts114k},
generates long-form traces for mathematics, science, code, and puzzle prompts
and verifies correctness before assembling the dataset. The corresponding
\textbf{OpenThinker} models use cross-entropy SFT on \textsc{Qwen2.5}
instruction-tuned backbones and are evaluated with \textsc{Evalchemy}. This
shared training and evaluation code supports comparisons between verified and
unverified trace sets. Later ablations report that answer verification helps at
32B but hurts at 7B. OpenThoughts3 omits answer
verification~\citep{guha2025openthoughts}.

\textbf{OpenThoughts2-1M} contains one million curated examples and is used to
train \textbf{OpenThinker2-32B}; the authors report matching
\textsc{DeepSeek-R1-Distill-32B} on \aime{} and
LiveCodeBench~\citep{openthoughts2_1m,openthinker2_32b}.
The \textbf{OpenThoughts3-1.2M} recipe was selected through more than 1,000
controlled experiments and uses \textsc{QwQ-32B} as its trace
teacher~\citep{openthoughts3_12m}. \textbf{OpenThinker3-7B} is trained by SFT
on the 1.2M-example mixture, reported as 850K mathematics, 250K code, and 100K
science examples, and its model card reports results on \aimefive,
LiveCodeBench, and GPQA Diamond~\citep{openthinker3_7b,openthoughts3_12m}.
The evaluation includes \textbf{OpenThinker2-32B} and
\textbf{OpenThinker3-1.5B}, trained with the OpenThoughts2 and OpenThoughts3
recipes, respectively.

\subsubsection{OpenAI gpt-oss-20b}

\textbf{gpt-oss-20b} is an MoE transformer post-trained with chain-of-thought
RL~\citep{openai_gptoss20b_hf,openai2025gptoss120bgptoss20bmodel}. It uses
MXFP4 quantization and has approximately 21B total parameters, with 3.6B active
per token. Its inference interface exposes three reasoning-effort settings:
\texttt{low}, \texttt{medium}, and \texttt{high}~\citep{oai_gptoss_api}.

\subsubsection{Microsoft Phi-4 reasoning models}

Microsoft's Phi-4 reasoning line specializes a 14B foundation model through
curated supervision and, for one variant, a short outcome-based RL stage.
Starting from \textsc{Phi-4}~\citep{abdin2024phi4},
\textbf{Phi-4-reasoning} applies SFT to prompts described as ``teachable''
(diverse and challenging but within the student's learnable regime), paired
with long-form demonstrations from \textsc{o3-mini}. The report studies this
prompt-selection criterion and evaluates longer reasoning budgets and repeated
sampling as inference controls~\citep{abdin2025phi4reasoning}.

\textbf{Phi-4-reasoning-plus} adds a short outcome-based RL stage that rewards
final-answer correctness on a verified mathematics set. The report associates
this stage with higher benchmark accuracy and longer average reasoning
traces~\citep{abdin2025phi4reasoning}. Microsoft also releases compact
variants. \textbf{Phi-4-mini-reasoning} (3.8B) uses mid-training on distilled
long-CoT data, long-CoT SFT, rollout-DPO preference optimization, and
verifiable-reward RL for mathematics~\citep{xu2025phi4minireasoning}.
\textbf{Phi-4-mini-flash-reasoning} uses a hybrid \textsc{SambaY} design and
Differential Attention for 64K-context reasoning~\citep{phi4miniflash_hf}. For
a 2K-token prompt and 32K-token generation under vLLM, the accompanying work
reports up to $10\times$ the decoding throughput of
Phi-4-mini-reasoning. It also reports higher scores on MATH-500, AIME 2024,
AIME 2025, and GPQA Diamond, although the Flash variant does not use RL
post-training~\citep{ren2025decoderhybriddecoder}.

\subsubsection{NVIDIA Nemotron reasoning models}

The evaluated NVIDIA releases include
\textbf{OpenReasoning-Nemotron-1.5B}, a dense Qwen2.5-derived SFT model for
mathematics, code, and science~\citep{openreasoning_nemotron_1_5b_hf}, and
\textbf{AceReason-Nemotron-1.1-7B}, which uses an SFT-and-RL pipeline for
mathematics and code~\citep{liu2025acereason}. Related NVIDIA work releases the
\textsc{OpenMathReasoning}~\citep{moshkov2025aimo2winningsolutionbuilding} and
\textsc{OpenCodeReasoning}~\citep{ahmad2025opencodereasoningiisimpletesttime,
ahmad2025opencodereasoning} data and model pipelines.

\textbf{NVIDIA-Nemotron-Nano-9B-v2} uses a hybrid Mamba--Transformer
architecture with controllable reasoning~\citep{nvidia2025nemotronnano2}. The
roster therefore contains a 1.5B dense baseline, a 7B SFT-and-RL specialist,
and a 9B hybrid model with reasoning controls.

\subsubsection{Open-R1 and OlympicCoder: open reproductions for math and competitive programming}

Open-R1~\citep{openr1_repo} publishes a staged reproduction of
\textsc{DeepSeek-R1}-style post-training, including its data mixtures,
verification procedures, and inference protocols.

A central component is
\textbf{Mixture-of-Thoughts}~\citep{mixture_of_thoughts}, a \(\sim\)350K
dataset of verified traces distilled from \textsc{DeepSeek-R1} across
mathematics, code, and science. \textbf{OpenR1-Distill-7B} is trained by SFT on
this mixture and provides a fixed checkpoint for analyzing repeated sampling,
aggregation, and subsequent verifiable-reward RL~\citep{openr1_distill_7b}.
Open-R1 also provides a mathematics-only corpus,
\textbf{OpenR1-Math-220k}, derived from NuminaMath 1.5 with multiple traces per
problem and automatic verification by \textsc{Math Verify}. When rule-based
checks are insufficient, \textsc{Llama-3.3-70B-Instruct} is the fallback judge;
the dataset card reports that this applies to 12\% of
samples~\citep{openr1_update2,openr1_math_220k}.

OlympicCoder applies the same approach to competitive programming, where
verification depends on test coverage. \textbf{CodeForces-CoTs} contains more
than 10K CodeForces problems and nearly 100K C++/Python traces generated by
\textsc{DeepSeek-R1}. The dataset is not exhaustively filtered; about 84\% of
its Python solutions pass the public tests~\citep{codeforces_cots}.
\textbf{OlympicCoder-7B} and \textbf{OlympicCoder-32B} fine-tune
\textsc{Qwen2.5-Coder} instruction models on decontaminated CodeForces-style CoT
planning data and are evaluated on an IOI'2024 subset, LiveCodeBench, and other
standardized competitive-programming benchmarks~\citep{olympiccoder7b,
olympiccoder32b,openr1_update3}. The mathematics corpus uses automatic
verification with an LLM fallback judge, whereas the code corpus relies on
public tests and is only partially verified.

\subsubsection{GAIR ``Less is More'': data-minimized reasoning and impact-aware RL}

GAIR's ``Less is More'' line tests whether data selection can substitute for
training-set volume. For \textbf{LIMO}, candidate solutions to 2,125
mathematics problems are sampled from \textsc{DeepSeek-R1},
\textsc{DeepSeek-R1-Distill-Qwen-32B}, and \textsc{QwQ-32B}. The highest-scoring
solution for each problem is retained, the resulting problem--solution pairs
are ranked, and the top 800 are used to fine-tune
\textsc{Qwen2.5-32B-Instruct}~\citep{ye2025limo}.
The study evaluates the resulting checkpoint with additional inference-time
compute, treating data selection and inference allocation as separate choices.
\textbf{LIMO-v2} pairs an updated 800-example dataset with a corresponding
\textsc{Qwen2.5-32B-Instruct} checkpoint
release~\citep{gair_limo_v2_modelcard,gair_limo_v2_dataset}.

\textbf{LIMR} introduces \emph{Learning Impact Measurement} (LIM), which
selects RL prompts by matching reward trajectories to the model's learning
dynamics. In its experiments, RL on 1,389 selected prompts matches or exceeds
RL on the full 8,523-prompt pool. The reported comparison also differs by model
scale: the smallest SFT sets work best at larger scales, whereas LIM-selected RL
improves the smaller models tested~\citep{li2025limr}.

\subsubsection{FuseAI/FuseO1: parameter-space merging}

Checkpoint merging combines trained models directly in parameter space without
gradient updates. FuseAI studies heterogeneous model
fusion~\citep{wan2025fusechat}.

FuseAI work includes distribution-level knowledge fusion~\citep{wan2024knowledge},
FuseChat's fuse-then-merge pipeline~\citep{wan2025fusechat}, and SFT+DPO
implicit fusion~\citep{yang2025fusechat}. FuseChat introduces \textbf{SCE}
(Select--Calculate--Erase), a parameter-matrix merge rule built around fusion
vectors (weight differences from a pivot) and a structured per-matrix
procedure~\citep{wan2025fusechat}. The \textbf{FuseO1} family combines
\textsc{DeepSeek-R1-Distill-Qwen-32B}, \textsc{QwQ-32B-Preview}, and
\textsc{Sky-T1-32B} variants through SCE, without gradient updates or new data
during the merge~\citep{fuseo1_repo,fuseo1_32b_preview,
fuseo1_flash_32b_preview}. Including a Flash parent is intended to combine long-
and short-reasoning behavior.

\subsubsection{Light-R1 and TinyR1: curriculum post-training and branch--merge distillation}

Light-R1 uses curriculum-style post-training to induce long-CoT behavior;
TinyR1 trains domain-specific branches and then merges them.

\textbf{Light-R1} uses two-stage curriculum SFT followed by semi-on-policy DPO
to train long-form reasoning and response style~\citep{wen2025lightr1}. In its
``from scratch'' setting, \textbf{Light-R1-32B} starts from
\textsc{Qwen2.5-32B-Instruct} and uses mathematics-focused
data~\citep{light_r1_32b_card}. A 3K ``stage-2'' hard long-CoT set is then
applied to \textsc{DeepSeek-R1-Distill-Qwen} checkpoints, producing
\textbf{Light-R1-7B-DS} and
\textbf{Light-R1-32B-DS}~\citep{light_r1_7b_ds_card,light_r1_32b_ds_card}.
The report attributes higher benchmark scores for these students to the
additional SFT stage. GRPO post-training produces \textbf{Light-R1-14B-DS}; the
report observes higher reward alongside longer
outputs~\citep{wen2025lightr1,light_r1_14b_ds_card}.

\textbf{TinyR1-32B-Preview} uses Branch--Merge Distillation. It distills
\textsc{DeepSeek-R1} into domain-specific branches through SFT and then merges
them. Relative to \textsc{DeepSeek-R1-Distill-Qwen-32B}, the paper reports
average gains of 5.5 points in mathematics, 4.4 in code, and 2.9 in science.
TinyR1 also approaches \textsc{DeepSeek-R1} on \aimefour{} under the reported
evaluation settings~\citep{sun2025tinyr1}.

\subsubsection{SimpleScaling s1.1: small-data SFT and budget forcing}

SimpleScaling's s1.1 family uses SFT on a small curated reasoning set and budget
forcing at inference time, replacing the earlier Gemini-generated traces with
DeepSeek-R1 traces. Budget forcing truncates or extends a reasoning trace to
control test-time compute~\citep{muennighoff2025s1}.

Across several Qwen model sizes, s1.1 combines sequential budget forcing with
parallel methods such as majority voting and REBASE; s1.1-32B is the primary
reference model~\citep{muennighoff2025s1}.

\subsubsection{LG EXAONE-4.0: unified chat and reasoning modes}

EXAONE-4.0 treats chat-style assistance and explicit reasoning as two modes of
one model family. Its unified post-training recipe jointly trains non-reasoning
and reasoning behavior, followed by preference-learning stages for correctness,
brevity, and language consistency in the 32B and 1.2B
models~\citep{exaone-4.0}.

The 32B model uses a 3:1 ratio of sliding-window to global-attention layers,
removes rotary positional embeddings from global-attention layers, and applies
QK-Reorder-Norm. Long-context post-training accompanies these architectural
changes. Post-training includes an RL
stage and multi-turn, long-horizon tool-use examples with execution
feedback~\citep{exaone-4.0}.

\section{Prompt and Inference Interfaces}
\label{app:prompt_templates}

Configurations use their model-native reasoning interfaces. Most receive a
zero-shot mathematics instruction requesting step-by-step reasoning and a
boxed final answer; gpt-oss also uses low, medium, and high effort controls.
Scoring uses the extracted final answer. In the repeated-mathematics block,
gpt-oss prompt coverage varies by task, as detailed in
\Cref{app:exp-math20}.

\section{Aggregation Reference Interface}
\label{app:aggregation_software}

The reference interface implements the aggregation protocol in
\Cref{ssec:aggregation} through three stages: evidence reducers map confidence
or reward signals to candidate scores; fixed-bank rules select an answer from
aligned answer and score arrays; and online predicates decide whether to stop
sampling or generation from the history observed so far. Separating these
stages allows the same candidate bank to be evaluated under several decision
rules and the same rule to be paired with different evidence sources, without
conflating candidate generation with post-generation selection.

The interface accepts a one-dimensional candidate bank for one problem or a
two-dimensional array for a batch. Score-free consensus, score-based selection,
and across-rollout stopping use the same extracted-answer representation.
Fixed-bank routines may also return the index of a representative candidate
when the selected answer needs a rationale. In
Listing~\ref{lst:aggregation_software}, plurality selects ``A'', whereas
Best-of-$N$ and softmax-weighted voting select ``B''; the causal stopping
predicate reads only the revealed answer prefix.

\begin{lstlisting}[
  style=pkgpython,
  escapeinside={(*@}{@*)},
  caption={Reference-interface pseudocode for fixed-bank aggregation and causal stopping. Higher candidate scores indicate stronger evidence.},
  label={lst:aggregation_software}
]
answers = ["A", "A", "A", "B"]
scores = [0.30, 0.30, 0.30, 0.85]

plurality = agg.majority_vote(answers)                  # "A"
best = agg.best_of_n(answers, scores)                   # "B"
weighted = agg.softmax_weighted_vote(
    answers, scores, temperature=0.1
)                                                       # "B"

observed = ["A"] * 8 + ["B"] * 2
stop, probability = agg.adaptive_consistency_stop(
    observed, threshold=0.95, return_prob=True
)                                                       # (True, 0.9673...)
\end{lstlisting}

\section{Data, Experiments, and Reproducibility}
\label{app:dataset}

We release 1,403,520 attempts across repeated mathematics, signal-rich
mathematics, and field-balanced \supergpqa{}, with 80 attempts per
model--question pair.
\Cref{tab:public-dataset-stats} shows that the 3,600-question field-balanced
\supergpqa{} collection accounts for most released attempts, whereas the two
mathematics collections cover 306 questions across nine tasks, so the aggregate
release count combines collections with very different question and task coverage. A \emph{response} is one model completion, and a \emph{bank} comprises the
responses from one model configuration on a fixed question set.

\begin{table}[!htbp]
\centering
\small
\setlength{\tabcolsep}{4pt}
\renewcommand{\arraystretch}{1.08}
\begin{tabular}{@{}lrrrr@{}}
\toprule
Collection & Model cfgs. & Tasks & Questions & Attempts \\
\midrule
Repeated mathematics & 20 & 4 & 120 & 192,000 \\
Signal-rich mathematics & 4 & 5 & 186 & 59,520 \\
Field-balanced \supergpqa{} & 4 & 1 & 3,600 & 1,152,000 \\
\midrule
\textbf{Total} & -- & -- & -- & \textbf{1,403,520} \\
\bottomrule
\end{tabular}
\caption{\textbf{Coverage of the public response collections.}}
\label{tab:public-dataset-stats}
\end{table}

\subsection{Data sources}\label{app:exp-sources}

Across the benchmark sources in \Cref{tab:dataset-sources}, question-set scale
ranges from 30-question competition sets to 12,032 \mmlupro{} questions; the
block-level estimates therefore rest on markedly different numbers of questions.

\begin{table}[H]
\centering
\footnotesize
\setlength{\tabcolsep}{3pt}
\renewcommand{\arraystretch}{1.08}
\begin{tabular}{@{}
  >{\raggedright\arraybackslash}p{0.20\linewidth}
  >{\raggedright\arraybackslash}p{0.60\linewidth}
  >{\raggedleft\arraybackslash}p{0.13\linewidth}
  @{}}
\toprule
Block & Source & Questions \\
\midrule
Broad & \mmlupro~\citep{mmlu_pro_dataset,wang2024mmlu}
  & 12,032 \\
Broad & \bbh{}~\citep{bbh_hf_dataset,suzgun2022challenging}
  & 6,511 \\
Repeated math & \aimefour~\citep{aime2024dataset}
  & 30 \\
Repeated math & \aimefive~\citep{aime2025dataset}
  & 30 \\
Repeated math & \hmmt~\citep{matharena_hmmt2025_dataset}
  & 30 \\
Repeated math & \brumo~\citep{matharena_brumo2025_dataset}
  & 30 \\
Signal-rich math & Five MathArena-hosted sets
  & 186 \\
Broad extension & \supergpqa~\citep{supergpqa_dataset,supergpqa2025}
  & 3,600 \\
\bottomrule
\end{tabular}
\caption{Question sources and sizes.}
\label{tab:dataset-sources}
\end{table}

The signal-rich study combines all 186 questions from \aimesix, \hmmtfebsix,
\hmmtnovfive, \cmimc, and \smt~\citep{matharena_aime2026_dataset,
matharena_hmmtfeb2026_dataset,matharena_hmmtnov2025_dataset,
matharena_cmimc2025_dataset,matharena_smt2025_dataset}. Its results use our
evaluation protocol.
\subsection{Metrics, reducers, and uncertainty}\label{app:exp-analysis}

For a question with \(N\) responses, let \(c\) be the number marked
correct by the benchmark adapter. The exact without-replacement coordinates
used throughout are
\begin{align}
\operatorname{Pass@}k
  &= 1 - \frac{\binom{N-c}{k}}{\binom{N}{k}},
&
\operatorname{pass}^{k}
  &= \frac{\binom{c}{k}}{\binom{N}{k}}.
\end{align}
The first asks whether a size-\(k\) subset contains at least one correct
candidate; the second asks whether every candidate in that subset is correct.
They describe a bank and do not specify which response a system submits.

We evaluate submitted-answer accuracy by replaying reducers on subsets of the
same bank. Budgets are \(k\in\{1,2,4,8,16,32,64,80\}\). A counter-derived
seed gives each question a reproducible random permutation, and prefixes form
nested uniform subsets without replacement. In the 20-configuration study we
enumerate all 80 singleton choices and use 200 paired subset replays for each
\(k>1\). The signal-rich mathematics study uses 2,000 replays for
\(1<k<80\), an exact average over all singletons at \(k=1\), and the full bank
at \(k=80\). Ties are resolved by the earliest candidate in the seeded random
order. A reducer's \(k=80\) value is therefore exact conditional on that fixed
tie-breaking order, not an average over every possible permutation.

The reported intervals resample questions after integrating over subset
replays. Experiment~2 uses 2,000 percentile bootstrap replicates; the
signal-rich mathematics analysis uses 10,000. These intervals are conditional
on the observed 80-response banks. They measure sensitivity to the question
composition, not the additional variability that would arise from generating
a new bank. Pooled summaries are item-weighted micro-averages. For
\supergpqa, field-balanced summaries give each of the 72 constructed fields
equal weight because every field contributes 50 questions.

The available signals constrain which analyses a bank supports:
\Cref{tab:reducer-inputs} maps each diagnostic or reducer to its required inputs
and distinguishes reference-free reducers from Compass Best-of-\(N\), which
requires a benchmark reference and is used only diagnostically.

\begin{table}[!htbp]
\centering
\small
\begin{tabularx}{\linewidth}{@{}lYY@{}}
\toprule
Reducer or statistic & Required signals & Interpretation \\
\midrule
Pass@\(k\), \(\mathrm{pass}^{k}\) & Binary outcome & Exact bank diagnostic \\
String plurality & Extracted answer & Deployable terminal reducer; literal strings \\
Sequence/mean log probability & Chosen-token log probabilities & Deployable confidence heuristic \\
Normalized-likelihood vote & Answer string and response likelihood & Deployable weighted terminal vote \\
Pointwise Best-of-\(N\) & Reference-free Qwen score & Deployable after paying verifier cost \\
Compass Best-of-\(N\) & Question, reference, and candidate & Reference-assisted diagnostic only \\
\bottomrule
\end{tabularx}
\caption{Inputs and roles of the reported analyses. ``Deployable'' here means
that the reducer does not require a benchmark reference answer; its compute is
still additional to candidate generation.}
\label{tab:reducer-inputs}
\end{table}

\subsection{Broad knowledge and symbolic reasoning}\label{app:exp-breadth}

\paragraph{Protocol.}
The broad block evaluates 27 model configurations on all 12,032 \mmlupro{}
and 6,511 \bbh{} questions~\citep{wang2024mmlu,suzgun2022challenging}.
\mmlupro{} uses final-option exact match, whereas \bbh{} uses its flexible
extractor.

Generation is zero-shot, with temperature 0.6, top-\(p=0.95\), frequency
penalty 0.1, a 32,768-token cap, and a fixed seed.

\paragraph{Response counts.}
The 27 configurations produce 500,661 responses across the two suites. This block lacks token-level
probabilities and structured finish reasons, so it supports neither
likelihood-based reducers nor reliable termination analysis.
The configuration orderings in \Cref{tab:breadth-results} differ between
\mmlupro{} and \bbh{}, so the suite-specific exact-match columns are retained
as separate descriptive results rather than pooled into one ranking.

\begin{table}[H]
\centering
\small
\begin{tabularx}{\linewidth}{@{}Yrr@{}}
\toprule
Configuration & \mmlupro & \bbh \\
\midrule
Bespoke-Stratos-32B & 67.37 & 75.41 \\
Bespoke-Stratos-7B & 46.68 & 51.50 \\
DeepSeek-R1-Distill-Llama-8B & 29.42 & 64.57 \\
DeepSeek-R1-Distill-Qwen-1.5B & 3.13 & 25.97 \\
DeepSeek-R1-Distill-Qwen-14B & 48.85 & 77.88 \\
DeepSeek-R1-Distill-Qwen-32B & 46.02 & 81.68 \\
DeepSeek-R1-Distill-Qwen-7B & 6.68 & 53.11 \\
FuseO1-DeepSeekR1-QwQ-32B-Preview & 45.05 & 81.22 \\
FuseO1-DeepSeekR1-QwQ-SkyT1-32B-Preview & 44.96 & 82.20 \\
FuseO1-DeepSeekR1-QwQ-SkyT1-Flash-32B-Preview & 44.52 & 81.45 \\
FuseO1-DeepSeekR1-Qwen2.5-Instruct-32B-Preview & 46.62 & 82.14 \\
LIMO & 65.61 & 74.41 \\
Light-R1-14B-DS & 38.23 & 74.09 \\
Light-R1-32B & 55.60 & 74.44 \\
Light-R1-7B-DS & 7.68 & 54.57 \\
OlympicCoder-32B & 50.78 & 57.79 \\
OlympicCoder-7B & 40.88 & 56.89 \\
OpenR1-Qwen-7B & 15.97 & 31.45 \\
OpenThinker-32B & 66.62 & 75.26 \\
OpenThinker-7B & 47.45 & 52.22 \\
QwQ-32B & 70.47 & 58.45 \\
Qwen2.5-32B-Instruct & 67.10 & 56.52 \\
Sky-T1-32B-Flash & 60.29 & 79.14 \\
Sky-T1-32B-Preview & 59.48 & 79.67 \\
Sky-T1-7B & 15.08 & 39.35 \\
TinyR1-32B-Preview & 58.32 & 77.39 \\
s1.1-32B & 64.23 & 41.48 \\
\bottomrule
\end{tabularx}
\caption{Broad-block exact match in percent. \bbh{} uses its flexible extractor.
The columns are separate descriptive results and are not averaged.}
\label{tab:breadth-results}
\end{table}

Empty \bbh{} extractions remain in the denominator and are scored incorrect.

\subsection{Twenty configurations on 2024--2025 mathematics}
\label{app:exp-math20}

\paragraph{Grid and models.}
The four sets are \aimefour, \aimefive, \hmmt, and \brumo,
with 30 questions each. \hmmt{} and \brumo{} are distributed through MathArena,
but the four-set combination is not itself a MathArena benchmark. Twenty
configurations generate 80 responses for each of 120 questions, yielding
192,000 responses.

\paragraph{Generation.}
Generation uses temperature 0.6, top-\(p=0.95\), one completion, and a
32,768-token cap. Realized-token log probabilities
and ranks are available at every generation position, but full next-token
distributions are not. These signals support sequence-likelihood,
mean-log-probability, and realized-rank summaries, but not entropy or other
full-distribution methods.
All non-gpt-oss configurations use the common boxed-answer instruction. Within
the three gpt-oss effort configurations, that instruction is present for 80\%
of the AIME'24 and AIME'25 attempts and absent from the BrUMO'25 and HMMT'25
attempts. Comparisons involving these configurations therefore combine model,
task, and prompt-interface differences.

\paragraph{CompassVerifier-7B.}
CompassVerifier-7B assigns an outcome score whose A/B/C labels mean correct,
incorrect, and invalid~\citep{CompassVerifier}; we use the probability assigned to A. Scores
or parsed answers that are unavailable are excluded without imputation.
Because the verifier sees the gold answer, we use it only as a
reference-assisted diagnostic.

Across the complete roster in \Cref{tab:exp2-roster-results}, Pass@80 exceeds
Acc.@1, the mean single-response accuracy, for every configuration; the
full-bank discovery statistic nevertheless does not define submitted-answer
accuracy.

\begin{table}[!t]
\centering
\small
\begin{tabularx}{\linewidth}{@{}Yrrr@{}}
\toprule
Configuration & Acc.@1 & Pass@8 & Pass@80 \\
\midrule
Qwen3-30B-A3B-Thinking-2507 & 75.56 & 85.83 & 91.67 \\
Qwen3-4B-Thinking-2507 & 66.55 & 80.40 & 87.50 \\
Phi-4-reasoning-plus & 65.56 & 84.19 & 91.67 \\
gpt-oss-20b high & 63.99 & 83.94 & 90.83 \\
gpt-oss-20b medium & 63.65 & 86.49 & 93.33 \\
AceReason-Nemotron-1.1-7B & 60.74 & 74.70 & 80.83 \\
Phi-4-reasoning & 59.97 & 82.81 & 90.83 \\
OpenThinker2-32B & 59.75 & 76.77 & 82.50 \\
FuseO1-DeepSeekR1-QwQ-SkyT1-Flash-32B-Preview & 58.52 & 74.05 & 81.67 \\
Light-R1-14B-DS & 58.27 & 74.09 & 81.67 \\
NVIDIA-Nemotron-Nano-9B-v2 & 54.72 & 72.12 & 79.17 \\
LIMO-v2 & 53.45 & 73.33 & 82.50 \\
gpt-oss-20b low & 47.39 & 72.39 & 85.00 \\
EXAONE-4.0-1.2B & 44.29 & 67.78 & 83.33 \\
OpenR1-Distill-7B & 43.68 & 65.66 & 78.33 \\
OpenThinker3-1.5B & 43.38 & 62.13 & 75.00 \\
OpenReasoning-Nemotron-1.5B & 40.73 & 62.07 & 78.33 \\
DeepSeek-R1-Distill-Qwen-1.5B & 25.30 & 47.47 & 64.17 \\
Sky-T1-32B-Flash & 25.27 & 43.95 & 55.83 \\
Bespoke-Stratos-7B & 18.39 & 36.22 & 54.17 \\
\bottomrule
\end{tabularx}
\caption{Complete fixed-roster results for the 2024--2025 mathematics block,
in percent. Acc.@1 averages the 80 response outcomes per question; Pass@\(k\)
is the exact without-replacement discovery statistic.}
\label{tab:exp2-roster-results}
\end{table}

\begin{table}[!htbp]
\centering
\small
\setlength{\tabcolsep}{4pt}
\begin{tabular}{@{}lrrr@{}}
\toprule
Qwen3-30B-A3B-Thinking-2507 & \(k=1\) & \(k=8\) & \(k=80\) \\
\midrule
Pass@\(k\) & 75.56 & 85.83 & 91.67 \\
\(\mathrm{pass}^{k}\) & 75.56 & 61.50 & 43.33 \\
String plurality & 75.56 & 77.74 & 78.33 \\
Sequence-log-probability Best-of-\(N\) & 75.56 & 78.45 & 77.50 \\
Mean-log-probability Best-of-\(N\) & 75.56 & 74.56 & 65.83 \\
Compass Best-of-\(N\) (reference-assisted) & 75.51 & 84.23 & 89.17 \\
\bottomrule
\end{tabular}
\caption{Selected repeated-mathematics results in percent, pooled over 120
questions. The \(k=1\) reducer values enumerate all singletons; \(k=8\) uses
200 paired subset replays; \(k=80\) is the full bank. Negative perplexity is
identical to mean-log-probability selection.}
\label{tab:exp2-selected}
\end{table}

For Qwen3-30B-A3B-Thinking-2507, \Cref{tab:exp2-selected} shows an
availability--selection gap at \(k=80\): Pass@80 is 91.67\%, whereas the best
deployable terminal reducer shown, string plurality, reaches 78.33\%.
The leading configuration's Acc.@1 has a 95\% prompt-bootstrap interval of
[68.81, 82.18]. Across the 20 configurations, the mean Spearman correlation
between subset and full-bank accuracy rankings is 0.967 at \(k=1\) and 0.992
at \(k=8\). This stability describes the present roster; it is not a general
sample-size guarantee.

\subsection{Signal-rich competition mathematics}\label{app:exp-matharena}

\paragraph{Generators and prompts.}
The generators are Qwen3.6-35B-A3B and
gpt-oss-20b~\citep{qwen36_hf,openai_gptoss20b_hf}.
Each receives the problem
in a zero-shot prompt that requests step-by-step reasoning and a boxed final
answer; no reference or evaluator signal is shown. Boxed answers are checked
for numeric or symbolic equivalence~\citep{evalscope_2024}.

Qwen uses temperature 1, top-\(p=0.95\), top-\(k=20\), and presence penalty
1.5. GPT-OSS uses temperature 1, top-\(p=1\), no top-\(k\) truncation, and no
presence penalty. Both use one completion and an 81,920-token generation cap.

\paragraph{Effective effort and sampling.}
Each model--effort condition contains \(186\times80=14{,}880\) responses. The
four analyzed banks are Qwen and gpt-oss at low, medium, and high effort. Three
additional gpt-oss runs used medium effort and serve only as diagnostics; they
are not pooled with the analyzed medium bank. Their Acc.@1 values range from
63.08\% to 63.47\%, compared with 62.80\% for the analyzed medium bank.

\paragraph{Token-level signals.}
We observe realized-token log probabilities and ranks and at most 20
alternatives per completion step. Ranks and probability mass beyond the first
20 alternatives are unavailable, and the recorded probabilities precede
decoding penalties and truncation.

\subsection{Verifier implementations and diagnostics}\label{app:exp-verifiers}

CompassVerifier scores final-answer outcomes with access to the reference
answer. The pointwise Qwen evaluator instead scores problem understanding,
reasoning validity, and conclusion support without a reference; its score is
used for reference-free Best-of-\(N\)
(\Cref{tab:verifier-comparison}).

\begin{table}[H]
\centering
\scriptsize
\setlength{\tabcolsep}{3pt}
\begin{tabularx}{\linewidth}{@{}p{0.15\linewidth}YY@{}}
\toprule
 & CompassVerifier-3B (OpenCompass) & Pointwise Qwen adaptation \\
\midrule
Input & Problem, reference answer, candidate response & Problem and one candidate response; no reference or other candidate \\
Judgment & A/B/C final-answer outcome; reasoning explicitly ignored & Problem understanding, reasoning validity, and conclusion support \\
Score & Direct A/B/C softmax; null-adjusted contextual A/B/C softmax & Expected A--T ordinal score per criterion, then mean \\
Role & Reference-assisted outcome diagnostic & Reference-free Best-of-\(N\) evidence \\
Main caveat & Gold access and neither distribution is calibrated & Ordinal, not calibrated; Qwen evaluates its own Qwen traces \\
\bottomrule
\end{tabularx}
\caption{The two verifier pipelines. The pointwise pipeline is inspired by
LLM-as-a-Verifier~\citep{kwok2026llmverifier} but is not its pairwise
tournament. CompassVerifier follows the released outcome-verifier prompt and
scoring formulation~\citep{CompassVerifier}.}
\label{tab:verifier-comparison}
\end{table}

\paragraph{CompassVerifier-3B (OpenCompass).}
The official prompt supplies the problem, standard answer, and candidate. It
asks for A (correct), B (incorrect), or C (invalid), instructs the verifier to
compare final answers, and explicitly says to ignore errors in the reasoning
when the final answer is correct~\citep{CompassVerifier}. From the label log
probabilities \((\ell_A,\ell_B,\ell_C)\), the direct score is
\[
p_{\mathrm{direct}}(z)
=\operatorname{softmax}_{z\in\{A,B,C\}}(\ell_z).
\]

The contextual variant subtracts a null-prompt baseline before normalizing;
let \(\ell_z^{\mathrm{slot}}\) and \(\ell_z^{\mathrm{null}}\) denote the label
log probabilities under the populated and null prompts:
\[
p_{\mathrm{ctx}}(z)
=\operatorname{softmax}_{z\in\{A,B,C\}}
  \left(\frac{\ell_z^{\mathrm{slot}}-\ell_z^{\mathrm{null}}}{1.5}\right).
\]
This adjustment is a diagnostic used by our pipeline; it is not a calibrated
correctness probability or the production label of CompassVerifier.

For inputs exceeding the context window, only the middle of the candidate is
shortened, preserving its prefix and suffix as well as the question and
reference.

\paragraph{Reference-free pointwise Qwen evaluator.}
The corpus-wide score adapts the expected-score idea of LLM-as-a-Verifier to
one candidate at a time~\citep{kwok2026llmverifier}. Qwen critiques problem
understanding, reasoning validity, and conclusion support, then assigns an
A--T ordinal distribution to each criterion. Unlike the cited pairwise method,
this evaluator ranks candidates directly without a tournament or
Bradley--Terry aggregation.

Let \(q_j(L)\) be the resulting distribution for criterion \(j\), with
\(v(A)=20,\ldots,v(T)=1\). We compute
\[
r_j=\frac{\sum_{L=A}^{T}q_j(L)v(L)-1}{19},
\qquad
s_{\mathrm{point}}=\frac{1}{3}\sum_{j=1}^{3}r_j.
\]
The score is in \([0,1]\), but the construction is ordinal rather than
probabilistic calibration, and low-mass letters may be censored by the returned
token probabilities. Qwen judging Qwen-generated responses is self-evaluation;
judging gpt-oss responses is cross-model evaluation. The reported selection
curves and intervals condition on the observed verifier scores and do not
include evaluator-rerun variability.

\begin{figure}[!htbp]
    \centering
    \includegraphics[width=\linewidth]{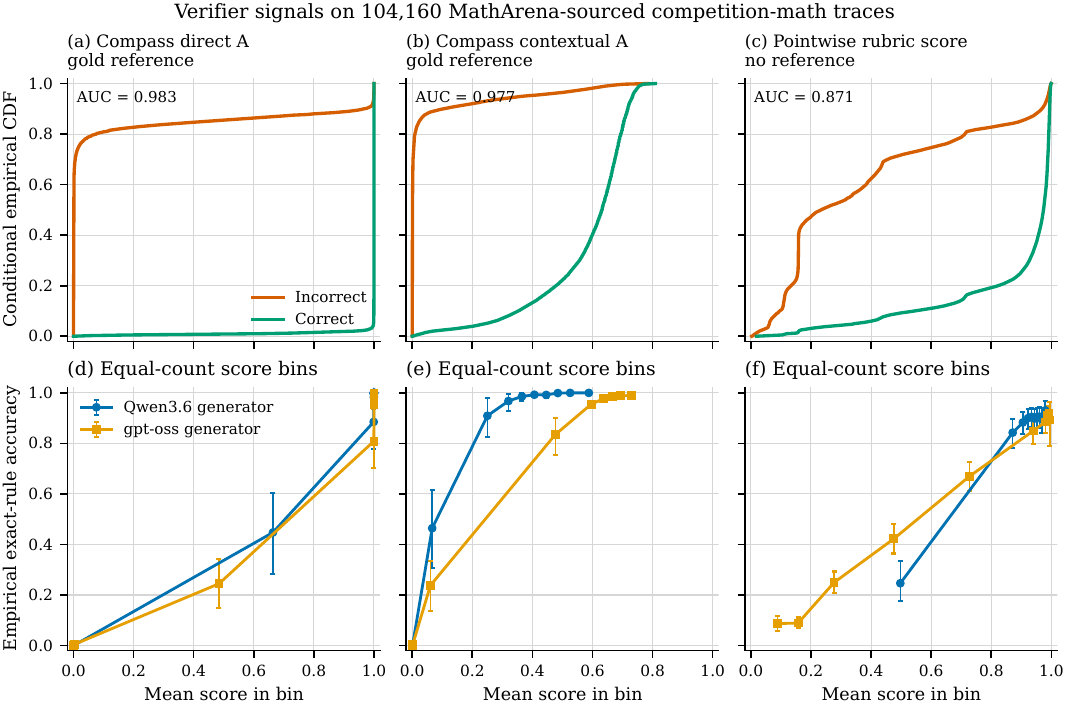}
    \caption{\textbf{Verifier signals against the independent rule-based
    outcome on 104,160 competition-math traces.} The figure compares direct
    and contextual Compass A scores with the reference-free pointwise Qwen
    score through conditional distributions and equal-count score bins. The
    lower panels separate Qwen-generated self-evaluations from gpt-oss
    cross-model evaluations. Compass sees the reference answer; pointwise Qwen
    does not. The binned empirical correctness curves describe ranking
    behavior on this fixed response set and should not be interpreted as
    calibration guarantees.}
    \label{fig:verifier-diagnostics}
\end{figure}

Against the rule-based outcome, the descriptive trace-level ROC AUC is 0.983
for direct Compass A, 0.977 for contextual Compass A, and 0.871 for the
pointwise score. The pointwise AUC is 0.744 on Qwen-generated traces and 0.876
on gpt-oss traces. This split does not identify a causal self-evaluation
effect because the generator families also differ in prompts, output
distributions, and base accuracy. A pooled AUC would conceal the observed
difference between generators.

\paragraph{Scaling results.}
\Cref{tab:matharena-results} reports the four canonical banks. String
plurality groups literal extracted strings rather than mathematically
equivalent forms. Pointwise Best-of-\(N\) uses only the reference-free score;
Compass scores are excluded from these deployed-selection
results.

\begin{table}[!htbp]
\centering
\small
\setlength{\tabcolsep}{3.5pt}
\begin{tabular}{@{}lrrrrrr@{}}
\toprule
Canonical bank & Acc.@1 & Pass@8 & Pass@80 & Str.-Plur.@80 & Pnt.-BoN@80 & Caps \\
\midrule
Qwen3.6 & 83.12 & 91.70 & 94.62 & 89.25 & 86.56 & 675 \\
gpt-oss low & 32.90 & 56.14 & 72.58 & 45.70 & 58.06 & 0 \\
gpt-oss medium & 62.80 & 85.99 & 91.94 & 80.11 & 75.81 & 32 \\
gpt-oss high & 73.03 & 88.12 & 93.55 & 85.48 & 81.72 & 2,399 \\
\bottomrule
\end{tabular}
\caption{Finite-bank and shared-bank results on the 186 MathArena-sourced
competition questions, in percent. Caps count responses that reach 81,920
tokens out of 14,880. Pnt.-BoN is pointwise Best-of-\(N\).}
\label{tab:matharena-results}
\end{table}

At \(k=80\), the 95\% prompt-bootstrap intervals for Pass@80 are
[90.86, 97.85], [66.13, 79.03], [87.63, 95.70], and [89.78, 96.77] in table
order. Pointwise Best-of-\(N\) intervals are [81.72, 91.40], [51.08, 65.05],
[69.35, 81.72], and [75.81, 87.10]. Several intervals overlap. Mean
completion counts per response are 34,382 tokens for Qwen and 1,781, 12,341,
and 37,264 for gpt-oss low, medium, and high. These costs exclude prompt tokens
and verifier computation. The high-effort cap rate is 16.12\%, so the effort
comparison is inseparable from the shared length limit.

\subsection{Field-balanced \supergpqa{} extension}\label{app:exp-supergpqa}

\supergpqa{} contains 26,529 multiple-choice questions across 285 graduate
disciplines~\citep{supergpqa2025}. Our 3,600-question sample includes 50 items
from each of 72 EvalScope fields, preventing large fields from dominating the
result~\citep{evalscope_2024}. The zero-shot prompt requests step-by-step reasoning and a final option
letter, which is extracted for binary scoring. The
four generation banks are Qwen3.6 and gpt-oss at low, medium, and high effort,
with 80 responses per question. Sampling and verification follow the signal-rich
mathematics protocol. The pointwise rubric is generalized to factual or
technical accuracy, logical coherence, and completeness. Together, the four
banks contain \(3{,}600\times4\times80=1{,}152{,}000\) responses.

\begin{figure}[H]
    \centering
    \includegraphics[width=\linewidth]{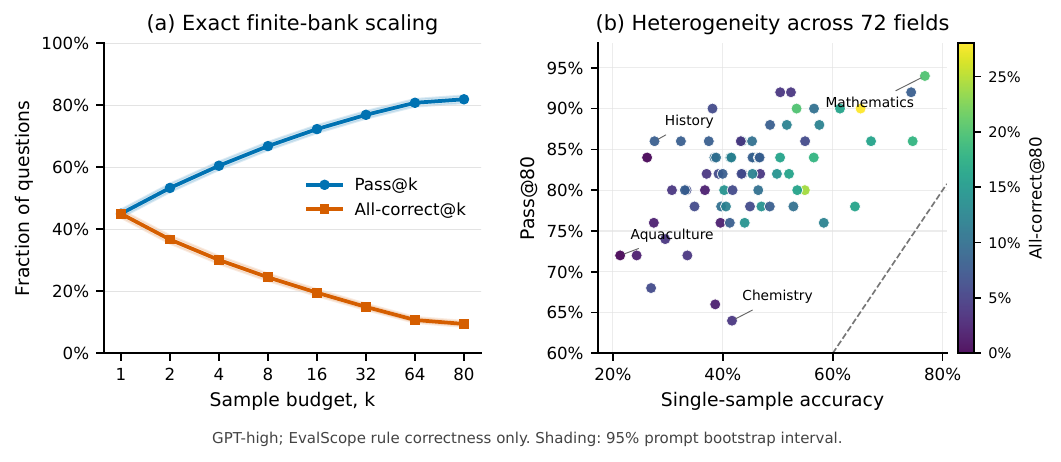}
    \caption{\textbf{\supergpqa{} scaling and field heterogeneity for the
    gpt-oss high bank.} The left panel gives exact finite-bank
    discovery and stability over 3,600 questions. In the right panel, each
    point is one of 72 equally sized fields: the axes compare single-response
    accuracy with Pass@80, and color shows the all-correct@80 coordinate. This
    figure uses only the rule-based correctness matrices; it does not
    substitute an LLM verifier for the benchmark outcome.}
    \label{fig:supergpqa-fields}
\end{figure}

For the gpt-oss high bank, the 288,000 responses have mean accuracy 45.03\%.
Exact Pass@80 is 81.94\%, while
\(\mathrm{pass}^{80}\) is 9.47\%. Of the 3,600 questions, 650 are never
answered correctly and 341 are answered correctly in all 80 responses.
Field-level response accuracy ranges from 21.33\% in Aquaculture to 76.78\%
in Mathematics. \Cref{fig:supergpqa-fields} reports finite-bank scaling and
field variation from the 3,600 by 80 binary matrix. These values are
descriptive of the equal-field construction, not estimates under \supergpqa{}'s
original field frequencies.

\end{document}